# A Feature Subset Selection Algorithm Automatic Recommendation Method

**Guangtao Wang**                                         GT.WANG@STU.XJTU.EDU.CN
**Qinbao Song**                                           QBSONG@MAIL.XJTU.EDU.CN
**Heli Sun**                                              HLSUN@MAIL.XJTU.EDU.CN
**Xueying Zhang**                                         ZHANGXUEYING.725@STU.XJTU.EDU.CN
*Department of Computer Science & Technology,*
*Xi'an Jiaotong University, 710049, China*

**Baowen Xu**                                             BWXU@NJU.EDU.CN
**Yuming Zhou**                                           ZHOUYUMING@NJU.EDU.CN
*Department of Computer Science & Technology,*
*Nanjing University, China*

## Abstract

Many feature subset selection (FSS) algorithms have been proposed, but not all of them are appropriate for a given feature selection problem. At the same time, so far there is rarely a good way to choose appropriate FSS algorithms for the problem at hand. Thus, FSS algorithm automatic recommendation is very important and practically useful. In this paper, a meta learning based FSS algorithm automatic recommendation method is presented. The proposed method first identifies the data sets that are most similar to the one at hand by the $k$-nearest neighbor classification algorithm, and the distances among these data sets are calculated based on the commonly-used data set characteristics. Then, it ranks all the candidate FSS algorithms according to their performance on these similar data sets, and chooses the algorithms with best performance as the appropriate ones. The performance of the candidate FSS algorithms is evaluated by a multi-criteria metric that takes into account not only the classification accuracy over the selected features, but also the runtime of feature selection and the number of selected features. The proposed recommendation method is extensively tested on 115 real world data sets with 22 well-known and frequently-used different FSS algorithms for five representative classifiers. The results show the effectiveness of our proposed FSS algorithm recommendation method.

## 1. Introduction

Feature subset selection (FSS) plays an important role in the fields of data mining and machine learning. A good FSS algorithm can effectively remove irrelevant and redundant features and take into account feature interaction. This not only leads up to an insight understanding of the data, but also improves the performance of a learner by enhancing the generalization capacity and the interpretability of the learning model (Pudil, Novovičová, Somol, & Vrňata, 1998a; Pudil, Novovičovà, Somol, & Vrňata, 1998b; Molina, Belanche, & Nebot, 2002; Guyon & Elisseeff, 2003; Saeys, Inza, & Larrañaga, 2007; Liu & Yu, 2005; Liu, Motoda, Setiono, & Zhao, 2010).

Although a large number of FSS algorithms have been proposed, there is no single algorithm which performs uniformly well on all feature selection problems. Experiments (Hall, 1999; Zhao & Liu, 2007) have confirmed that there could exist significant differences of performance (e.g., classification accuracy) among different FSS algorithms over a given data set. That means for a given data set, some FSS algorithms outperform others.

This raises a practical and very important question: which FSS algorithms should be picked up for a given data set? The common solution is to apply all candidate FSS algorithms to the given data set, and choose one with the best performance by the cross-validation strategy. However, this solution is quite time-consuming especially for high-dimensional data (Brodley, 1993).

For the purpose of addressing this problem in a more efficient way, in this paper, an FSS algorithm automatic recommendation method is proposed. The assumption underlying our proposed method is that the performance of an FSS algorithm on a data set is related to the characteristics of the data set. The rationality of this assumption can be demonstrated as follows:

1) Generally, when a new FSS algorithm is proposed, its performance needs to be extensively evaluated at least on real world data sets. However, the published FSS algorithms are rarely tested on the identical group of data sets (Hall, 1999; Zhao & Liu, 2007; Yu & Liu, 2003; Dash & Liu, 2003; Kononenko, 1994). That is, for any two algorithms, they are usually tested on the different data. This implies that the performance of an FSS algorithm biases to some data sets.

2) At the same time, the famous NFL (No Free Lunch) (Wolpert, 2001) theory tells us that, for a particular data set, different algorithms have different data-conditioned performance, and the performance differences vary with data sets.

The above evidences imply that there is a relationship between the performance of an FSS algorithm and the characteristics of data sets. In this paper, we intend to explore this relationship and utilize it to automatically recommend appropriate FSS algorithm(s) for a given data set. The recommendation process can be viewed as a specific application of meta-learning (Vilalta & Drissi, 2002; Brazdil, Carrier, Soares, & Vilalta, 2008) that has been used to recommend algorithms for classification problems (Ali & Smith, 2006; King, Feng, & Sutherland, 1995; Brazdil, Soares, & Da Costa, 2003; Kalousis, Gama, & Hilario, 2004; Smith-Miles, 2008; Song, Wang, & Wang, 2012).

To model this relationship, there are three fundamental issues to be considered: i) which features (often are referred to as meta-features) are used to characterize a data set; ii) how to evaluate the performance of an FSS algorithm and identify the applicable one(s) for a given data set; iii) how to recommend FSS algorithm(s) for a new data set.

In this paper, the meta-features, which are frequently used in meta-learning (Vilalta & Drissi, 2002; Ali & Smith, 2006; King et al., 1995; Brazdil et al., 2003; Castiello, Castellano, & Fanelli, 2005), are employed to characterize data sets. At the same time, a multi-criteria metric, which takes into account not only the classification accuracy of a classifier with an FSS algorithm but also the runtime of feature selection and the number of selected features, is used to evaluate the performance of the FSS algorithm. Meanwhile, a $k$-NN ($k$-Nearest Neighbor) based method is proposed to recommend FSS algorithm(s) for a new data set.

Our proposed FSS algorithm recommendation method has been extensively tested on 115 real world data sets with 22 well-known and frequently-used different FSS algorithms for five representative classifiers. The results show the effectiveness of our proposed recommendation method.

The rest of this paper is organized as follows. Section 2 introduces the preliminaries. Section 3 describes our proposed FSS algorithm recommendation method. Section 4 provides the experimental results. Section 5 conducts the sensitivity analysis of the number of the nearest data sets on the recommendation results. Finally, section 6 summarizes the work and draws some conclusions.

## 2. Preliminaries

In this section, we first describe the meta-features used to characterize data sets. Then, we introduce the multi-criteria evaluation metric used to measure the performance of FSS algorithms.

### 2.1 Meta-features of Data Sets

Our proposed FSS algorithm recommendation method is based on the relationship between the performance of FSS algorithms and the meta-features of data sets.

The recommendation can be viewed as a data mining problem, where the performance of FSS algorithms and the meta-features are the target function and the input variables, respectively. Due to the ubiquity of "Garbage In, Garbage Out" (Lee, Lu, Ling, & Ko, 1999) in the field of data mining, the selection of the meta-features is crucial for our proposed FSS recommendation method.

The meta-features are measures that are extracted from data sets and can be used to uniformly characterize different data sets, where the underlying properties are reflected. The meta-features should be not only conveniently and efficiently calculated, but also related to the performance of machine learning algorithms (Castiello et al., 2005).

There has been 15 years of research to study and improve on the meta-features proposed in the StatLog project (Michie, Spiegelhalter, & Taylor, 1994). A number of meta-features have been employed to characterize data sets (Brazdil et al., 2003; Castiello et al., 2005; Michie et al., 1994; Engels & Theusinger, 1998; Gama & Brazdil, 1995; Lindner & Studer, 1999; Sohn, 1999), and have been demonstrated working well in modeling the relationship between the characteristics of data sets and the performance (e.g., classification accuracy) of learning algorithms (Ali & Smith, 2006; King et al., 1995; Brazdil et al., 2003; Kalousis et al., 2004; Smith-Miles, 2008). As these meta-features do characterize data sets themselves, and have no connection with learning algorithms and their types, so we use them to model the relationship between data sets and FSS algorithms.

The most commonly used meta-features are established focusing on the following three aspects of a data set: i) general properties, ii) statistic-based properties, and iii) information-theoretic based properties (Castiello et al., 2005). Table 1[1] shows the details.

---

1. In order to compute the information-theoretic features, for data sets with continuous-valued features, if needed, the well-known MDL (Minimum Description Length) method with Fayyad & Irani criterion was used to discretize the continuous values.

| Category | Notation | Measure description |
|---|---|---|
| General properties | $I$ | Number of instances |
| | $F$ | Number of features |
| | $T$ | Number of target concept values |
| | $D$ | Data set dimensionality, $D = I/F$ |
| Statistical based properties | $\overline{\rho}(X,Y)$ | Mean absolute linear correlation coefficient of all possible pairs of features |
| | $\overline{Skew}(X)$ | Mean skewness |
| | $\overline{Kurt}(X)$ | Mean kurtosis |
| Information-theoretic properties | $H(C)_{norm}$ | Normalized class entropy |
| | $\overline{H}(X)_{norm}$ | Mean normalized feature entropy |
| | $\overline{MI}(C,X)$ | Mean mutual information of class and attribute |
| | $MI(C,X)_{max}$ | Maximum mutual information of class and attribute |
| | $EN_{attr}$ | Equivalent number of features, $EN_{attr} = H(C)/\overline{MI}(C,X)$ |
| | $NS_{ratio}$ | Noise-signal ratio, $NS_{ratio} = (\overline{H}(X) - \overline{MI}(C,X))/\overline{MI}(C,X)$ |

Table 1: Meta-features used to characterize a data set

## 2.2 Multi-criteria Metric for FSS Algorithm Evaluation

In this section, first, the classical metrics evaluating the performance of FSS algorithm are introduced. Then, by analyzing the user requirements in practice application, based on these metrics, a new and user-oriented multi-criteria metric is proposed for FSS algorithm evaluation by combining these metrics together.

### 2.2.1 Classical Performance Metrics

When evaluating the performance of an FSS algorithm, the following three metrics are extensively used in feature selection literature: i) classification accuracy , ii) runtime of feature selection, and iii) number of selected features.

1) The *classification accuracy* (*acc*) of a classifier with the selected features can be used to measure how well the selected features describe a classification problem. This is because for a given data set, different feature subsets generally result in different classification accuracies. Thus, it is reasonable that the feature subset with higher classification accuracy has stronger capability of depicting the classification problem. The classification accuracy also reflects the ability of an FSS algorithm in identifying the salient features for learning.

2) The *runtime* (*t*) of feature selection measures the efficiency of an FSS algorithm for picking up the useful features. It is also viewed as a metric to measure the cost of feature selection. The longer the runtime, the higher the expenditure of feature selection.

3) The *number of selected features* (*n*) measures the simplicity of the feature selection results. If the classification accuracies with two FSS algorithms are similar, we usually favor the algorithm with fewer features.

Feature subset selection aims to improve the performance of learning algorithms which usually is measured with classification accuracy. The FSS algorithms with higher classification accuracy are in favor. However, this does not mean that the runtime and the number of selected features could be ignored. This can be explained by the following two considerations:

1) Suppose there are two different FSS algorithms $A_i$ and $A_j$, and a given data set $D$. If the classification accuracy with $A_i$ on $D$ is slightly greater than that with $A_j$, but the runtime of $A_i$ and the number of features selected by $A_i$ are much greater than those of $A_j$, then $A_j$ is often chosen.

2) Usually, we do not prefer to use the algorithms with higher accuracy but longer runtime, so is those with lower accuracy but shorter runtime. Therefore, we need a tradeoff between classification accuracy and the runtime of feature selection/the number of selected features. For example, in real-time systems, it is impossible to choose the algorithm with high time-consumption even if its classification accuracy is high.

Therefore, it is necessary to allow users making a user-oriented performance evaluation for different FSS algorithms. For this purpose, it is needed to address the problem of how to integrate classification accuracy with the runtime of feature selection and the number of selected features to obtain a unified metric. In this paper, we resort to the multi-criteria metric to explore this problem. The underlying reason lies that the multi-criteria metric has been successfully used to evaluate data mining algorithms by considering the positive properties (e.g. classification accuracy) and the negative ones (e.g. runtime and number of selected features) simultaneously (Nakhaeizadeh & Schnabl, 1997, 1998).

When comparing two algorithms, besides the metrics used to evaluate their performance, the *ratio* of the metric values can also be used. For example, suppose $A_1$ and $A_2$ are two different FSS algorithms, if $A_1$ is better than $A_2$ in terms of classification accuracy, i.e., $acc_1 > acc_2{}^2$, then ratio $acc_1/acc_2 > 1$ can be used to show $A_1$ is better than $A_2$ as well. On the contrary, for the negative metrics runtime of feature selection and number of of the selected features, the corresponding ratio $< 1$ means a better algorithm.

Actually, a multi-criteria metric *adjusted ratio of ratios* ($ARR$) (Brazdil et al., 2003), which combines classification accuracy and runtime together as a unified metric, has been proposed to evaluate the performance of a learning algorithm. We extend $ARR$ by integrating it with the runtime of feature selection and the number of selected features, so a new multi-criteria metric $EARR$ (*extened ARR*) is proposed. In the following discussion, we will show that the new metric $EARR$ is more inclusive, very flexible, and easy to understand.

### 2.2.2 Multi-Criteria Metric EARR

Let DSet $= \{D_1, D_2, \cdots, D_N\}$ be a set of $N$ data sets, and ASet $= \{A_1, A_2, \cdots, A_M\}$ be a set of $M$ FSS algorithms. Suppose $acc_i^j$ is the classification accuracy of a classifier with FSS algorithm $A_i$ on data set $D_j$ ($1 \leq i \leq M, 1 \leq j \leq N$), and $t_i^j$ and $n_i^j$ denote the runtime and the number of selected features of FSS algorithm $A_i$ on data set $D_j$, respectively. Then the $EARR$ of $A_i$ to $A_j$ over $D_k$ is defined as

$$EARR_{A_i, A_j}^{D_k} = \frac{acc_i^k/acc_j^k}{1 + \alpha \cdot \log{(t_i^k/t_j^k)} + \beta \cdot \log{(n_i^k/n_j^k)}} \quad (1 \leq i \neq j \leq M, 1 \leq k \leq N), \qquad (1)$$

where $\alpha$ and $\beta$ are the user-predefined parameters which denote the relative importance of the runtime of feature selection and the number of selected features, respectively.

The computation of the metric $EARR$ is based on the ratios of the classical FSS algorithm performance metrics, the classification accuracy, the runtime of feature selection and the number of selected features. From the definition we can know that $EARR$ evaluates an FSS algorithm by comparing it with another algorithm. This is reasonable since it is more objective to assert an algorithm is good or not by comparing it with another one instead of just focusing on its own performance. For example, suppose there is a classifier with 70% classification accuracy on a data set, we will get confused on whether the classifier is good or not. However, if we compare it with another classifier that can obtain 90% classification accuracy on the same data set, then we can definitely say that the first classifier is not good compared with the second one.

---

2. Where $acc_1$ and $acc_2$ are the corresponding classification accuracies of algorithms $A_1$ and $A_2$, respectively.

Noted that, in practice, the runtime difference between two different FSS algorithms usually can be quite great. Meanwhile, for high-dimensional data sets, the difference of the number of selected features for two different FSS algorithms can be great as well. Thus, the ratio of runtime and the ratio of the number of selected features usually have much more wide ranges than that of the classification accuracy. If the simple ratio of runtime and the simple ratio of the number of selected features are employed, they would dominate the value of $EARR$, and the ratio of the classification accuracy will be drowned. In order to avoid this situation, the common logarithm (i.e., the logarithm with base 10) of the ratio of runtime and the common logarithm of the ratio of the number of selected features are employed.

The parameters $\alpha$ and $\beta$ represent the amount of classification accuracy that a user is willing to trade for a 10 times speedup/reduction on the runtime of feature selection/the number of selected features, respectively. This allows users to choose the algorithms with shorter runtime and less features but acceptable accuracy. This can be illustrated by the following example. Suppose that $acc_i^k = (1 + \alpha + \beta) \cdot acc_j^k$, the runtime of algorithm $A_i$ on a given data set is 10 times of that of $A_j$ (i.e., $t_i^k = 10 \cdot t_j^k$), and the number of selected features of algorithm $A_i$ is 10 times of that of $A_j$ (i.e., $n_i^k = 10 \cdot n_j^k$). Then, according to Eq. (1), $EARR_{A_i,A_j}^{D_k} = 1$, and $EARR_{A_j,A_i}^{D_k} = \frac{1}{1-(\alpha+\beta)^2} > 1$.[3] In this case, $A_j$ outperforms $A_i$. If a user prefers fast algorithms with less features, $A_j$ will be his/her choice.

The value of $EARR$ varies around 1. The value of $EARR_{A_i,A_j}^{D_k}$ is greater than (or equal to, or smaller than) that of $EARR_{A_j,A_i}^{D_k}$ indicates that $A_i$ is better than (or equal to, or worse than) $A_j$.

Eq. (1) can be directly used to evaluate the performance of two different FSS algorithms. When comparing multiple FSS algorithms, the performance of any algorithm $A_i \in$ ASet on a given data set $D$ can be evaluated by the metric $EARR_{A_i}^D$ defined as follows:

$$EARR_{A_i}^D = \frac{1}{M-1} \sum_{j=1 \wedge j \neq i}^{M} EARR_{A_i,A_j}^D. \tag{2}$$

This equation shows that the $EARR$ of an FSS algorithm $A_i$ on $D$ is the arithmetic mean of the $EARR_{A_i,A_j}^D$ of $A_i$ to other algorithm $A_j$ on $D$. That is, the performance of any FSS algorithm $A_i \in$ ASet is evaluated based on the comparisons with other algorithms in $\{$ASet $-\{A_i\}\}$. The larger the value of $EARR$, the better the corresponding FSS algorithm on the given data set $D$.

---

3. Since $\log(x/y) = -\log(y/x)$ and $(\alpha + \beta)^2 > 0$

## 3. FSS Algorithm Recommendation Method

In this section, we first give the framework of the FSS algorithm recommendation. Then, we introduce the nearest neighbor based recommendation method in detail.

### 3.1 Framework

Based on the assumption that there is a relationship between the performance of an FSS algorithm on a given data set and the data set characteristics (aka meta-features), our proposed recommendation method firstly constructs a meta-knowledge database consisting of data set meta-features and FSS algorithm performance. After that, with the help of the meta-knowledge database, a $k$-NN based method is used to model this relationship and recommend appropriate FSS algorithms for a new data set.

Therefore, our proposed recommendation method consists of two parts: *Meta-knowledge Database Construction* and *FSS Algorithm Recommendation*. Fig. 1 shows the details.

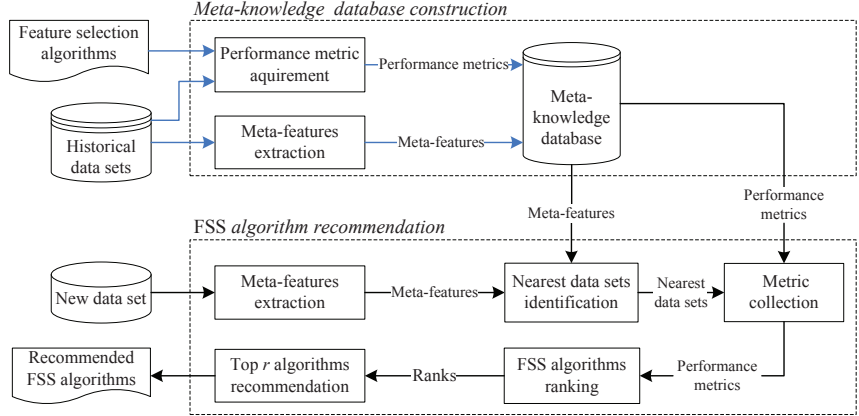

Figure 1: Framework of feature subset selection algorithm recommendation

1) *Meta-Knowledge Database Construction*

As mentioned previously, the meta-knowledge database consists of the meta-features of a set of historical data sets and the performance of a group of FSS algorithms on them. It is the foundation of our proposed recommendation method, and the effectiveness of the recommendations depends heavily on this database.

The meta-knowledge database is constructed by the following three steps. Firstly, the meta-features in Table 1 are extracted from each historical data set by the module "Meta-features extraction". Then, each candidate FSS algorithm is applied on each historical data set. The classification accuracy, the runtime of feature selection and the number of selected features are recorded, and the corresponding value of the performance metric $EARR$ is calculated. This is accomplished by the module "Performance metric calculation". Finally, for each data set, a tuple, which is composed of the meta-features and the values of the performance metric $EARR$ for all the candidate FSS algorithms, is obtained and added into the knowledge database.

2) *FSS Algorithm Recommendation*

Based on the introduction of the first part "Meta-knowledge Database Construction" we presented above, the learning target of the meta-knowledge data is a set of $EARR$ values instead of an appropriate FSS algorithm. In this case, it has been demonstrated that the researchers usually resort to the instance-based or $k$-NN (nearest neighbors) methods or their variations (Brazdil et al., 2003, 2008) for algorithm recommendation. Thus, a $k$-NN based FSS algorithm recommendation procedure is proposed.

When recommending FSS algorithms for a new data set, firstly, the meta-features of this data set are extracted. Then, the distance between the new data set and each historical data set is calculated according to the meta-features. After that, its $k$ nearest data sets are identified, and the $EARR$ values of the candidate FSS algorithms on these $k$ data sets are retrieved from the meta-knowledge database. Finally, all the candidate FSS algorithms are ranked according to these $EARR$ values, where the algorithm with the highest $EARR$ achieves the top rank, the one with the second highest $EARR$ gets second rank, and so forth, and the top $r$ algorithms are recommended.

## 3.2 Recommendation Method

To recommend appropriate FSS algorithms for a new data set $D_{new}$ based on its $k$ nearest data sets, there are two foundational issues to be solved: i) how to identify its $k$ nearest data sets, and ii) how to recommend appropriate FSS algorithms based on these $k$ data sets.

1) *k nearest data sets identification*

The $k$ nearest data sets of $D_{new}$ are identified by calculating the distance between $D_{new}$ and each historical data set based on their meta-features. The smaller the distance, the more similar the corresponding data set to $D_{new}$.

In order to effectively calculate the distance between two data sets, the $L_1$ norm distance (Atkeson, Moore, & Schaal, 1997) is adopted since it is easy to understand and calculate, and its ability in measuring the similarity between two data sets has been demonstrated by Brazdil et al. (2003).

Let $F_i = <f_{i,1}, f_{i,2}, \cdots, f_{i,h}>$ be the meta-features of data set $D_i$, where $f_{i,p}$ is the value of $p$th feature of $F_i$ and $h$ is the length of the meta-features. The $L_1$ norm distance between data sets $D_i$ and $D_j$ can be formulated as

$$dist(D_i, D_j) = \|F_i - F_j\|_1 = \sum_{p=1}^{h} |f_{i,p} - f_{j,p}|. \tag{3}$$

It is worth noting that the ranges of different meta-features are quite different. For example, of the meta-features introduced in Table 1, the value of normalized class entropy varies from 0 to 1, while the number of instances can be millions. Thus, if these meta-features with different ranges are directly used to calculate the distance between two data sets, the meta-features with large range would dominate the distance, and the meta-features with small range will be ignored. In order to avoid this problem, the *0-1 standardized method* (Eq. (4)) is employed to make all the meta-features have the same range $[0, 1]$.

$$\frac{f_{i,p} - \min(f_{\cdot,p})}{\max(f_{\cdot,p}) - \min(f_{\cdot,p})}, \tag{4}$$

where $f_{i,p}$ $(1 \leq p \leq h)$ is the value of the $p$th meta-feature of data set $D_i$, $\min(f_{.,p})$ and $\max(f_{.,p})$ denote the minimum and maximum values of the $p$th meta-feature over historical data sets, respectively.

2) *FSS algorithm recommendation*

Once getting the $k$ nearest data sets of $D_{new}$, the performance of the candidate FSS algorithms on $D_{new}$ is evaluated according to those on the $k$ nearest data sets. Then, the algorithms with the best performance are recommended.

Let $\mathrm{D}_{knn} = \{D_1, D_2, \cdots, D_k\}$ be the $k$ nearest data sets of $D_{new}$ and $EARR_{A_i}^{D_j}$ be the performance metric of the FSS algorithm $A_i$ on data set $D_j \in \mathrm{D}_{knn}$ $(1 \leq j \leq k)$. Then the performance of $A_i$ on the new data set $D_{new}$ can be evaluated by

$$EARR_{A_i}^{\mathrm{D}_{knn}} = \sum_{j=1}^{k} \omega_j \cdot EARR_{A_i}^{D_j}, \text{ where } \omega_j = d_j^{-1} / \sum_{t=1}^{k} d_t^{-1}, \ d_j = dist(D_{new}, D_j). \quad (5)$$

Eq. (5) indicates that the performance of the FSS algorithm $A_i$ on $D_{new}$ is evaluated by its performance on the $\mathrm{D}_{knn}$ of $D_{new}$. For a data set $D_j \in \mathrm{D}_{knn}$, the smaller the distance $d_j$ between itself and $D_{new}$, the more similar the two data sets. This means for two data sets $D_p$ and $D_q$, if $d_p < d_q$ then the data set $D_p$ is more similar to $D_{new}$, so the $EARR$ of $A_i$ on $D_p$ is more important for evaluating the performance of $A_i$ on $D_{new}$. Thus, the weighted average, which takes into account the relative importance of each data set in $\mathrm{D}_{knn}$ rather than treating each data set equally, is employed. Moreover, in the domain of machine learning, the reciprocal of the distance usually is used to measure the similarity. So the $\omega_j = d_j^{-1} / \sum_{t=1}^{k} d_t^{-1}$ is used as the weight of the $EARR$ of $A_i$ on $D_j \in \mathrm{D}_{knn}$.

According to the $EARR$ of each candidate FSS algorithm in ASet on $D_{new}$, a rank of these candidate FSS algorithms can be obtained. The greater the $EARR$, the higher the rank. Then, the top $r$ (e.g., $r = 3$ in this paper) FSS algorithms are picked up as the appropriate ones for the new data set $D_{new}$.

Procedure *FSSAlgorithmRecommendation* shows the pseudo-code of the recommendation.

*Time complexity.* The recommendation procedure consists of two parts. In the first part (lines 1-3), the $k$ nearest data sets of the given new data set $D$ are identified. Firstly, the meta-features $F$ of $D$ are extracted by function MetaFeatureExtraction(). Then, the $k$-nearest historical data sets are identified by function NeighborRecognition() based on the distance between $F$ and the meta-features $F_i$ of each historical data set $D_i$. Suppose that $P$ is the number of instances and $Q$ is the number of features in the given data set $D$, the time complexity of function MetaFeatureExtraction() is $O(P + Q)$. For function NeighborRecognition(), the time complexity is $O(n)$ which depends on the number of the historical data sets $n$. Consequently, the time complexity of the first part is $O(P+Q)+O(n)$.

In the second part (lines 4-8), the $r$ FSS algorithms are recommended for the data set $D$. Since the weights and $EARR$s of the $k$ nearest data sets can be obtained directly, the time complexity of these two steps is $O(1)$. The time complexity for estimating and ranking the $EARR$s of the algorithms in ASet is $O(k \cdot m) + O(m \cdot \log(m))$, where $k$ is the preassigned number of the nearest data sets and $m$ is the number of the candidate FSS algorithms.

---

**Procedure FSSAlgorithmRecommendation**

---

   **Inputs** :

        $D$ - a new given data set;

        $DSet$ - $\{D_1, D_2, \cdots, D_n\}$, historical data sets;

        $ASet$ - $\{A_1, A_2, \cdots, A_m\}$, candidate FSS algorithms;

        $MetaDataBase$ - $\{<F_i, EARRs_i>|1 \leq i \leq n\}$ where $F_i$ and $EARRs_i$ are the
meta-features and the $EARRs$ of $ASet$ on $D_i$, respectively;

        $k$ - the predefined number of the nearest neighbors;

        $r$ - the number of recommended FSS algorithms.

   **Output**: $RecAlgs$ - Recommended FSS algorithms for $D$

    //Part 1: Recognition of the $k$ nearest data sets for $D$

**1**   F = `MetaFeatureExtraction` $(D)$;//Extract meta-features from $D$

**2**   $MetaFeatureSet = \{F_1, F_2, \cdots, F_n\}$;//Meta-feature of each data set in $DSet$

**3**   $Neighbors =$ `NeighborRecognition` $(k, F, MetaFeatureSet)$;

    //Part 2: Appropriate FSS algorithm recommendation

**4**   $WeightSet =$ calculate the weight for each data set in $Neighbors$ //See Eq. (5)

**5**   $EARRSet =$ the corresponding $EARRs$ for each data set in $Neighbors$ from $MetaDataBase$;

**6**   Estimate the $EARR$ of each FSS algorithm $\in ASet$ on $D$ according to $WeightSet$ and $EARRSet$
    by Eq. (5) and rank the algorithms in $ASet$ based on these $EARRs$;

**7**   $RecAlgs =$ top $r$ FSS algorithms;

**8**   **return** $RecAlgs$;

---

To sum up, the time complexity of the recommendation procedure is $O(P+Q)+O(n)+O(k \cdot m)+O(m \cdot \log(m))$. In practice, for a data set $D$ that needs to conduct feature selection, the number of the instances $P$ and/or the number of the features $Q$ in $D$ are much greater than the number of the nearest data sets $k$ and the number of the candidate FSS algorithms $m$, so the major time consumption of this recommendation procedure is determined by the first part.

## 4. Experimental Results and Analysis

In this section, we experimentally evaluate the proposed feature subset selection (FSS) algorithm recommendation method by recommending algorithms over the benchmark data sets.

### 4.1 Benchmark Data Sets

115 extensively-used real world data sets, which come from different areas such as Computer, Image, Life, Biology, Physical and Text [4], are employed in our experiment. The sizes of these data sets vary from 10 to 24863 instances, and the numbers of their features are between 5 and 27680.

The statistical summary of these data sets is shown in Table 2 in terms of the number of instances (denoted as I), the number of features (denoted as F) and the number of target concepts (denoted as T).

---

4. These data sets are available from `http://archive.ics.uci.edu/ml/datasets.html`, `http://featureselection.asu.edu/datasets.php`, `http://sci2s.ugr.es/keel/datasets.php`, `http://www.upo.es/eps/bigs/datasets.html`, and `http://tunedit.org/repo/Data`, respectively.

| Data ID | Data Name | I | F | T | Data ID | Data Name | I | F | T |
|---|---|---|---|---|---|---|---|---|---|
| 1 | ada_agnostic | 4562 | 49 | 2 | 59 | Lymphoma96x4026+9classes | 96 | 4027 | 9 |
| 2 | ada_prior | 4562 | 15 | 2 | 60 | mfeat-fourier | 2000 | 77 | 10 |
| 3 | anneal | 898 | 39 | 6 | 61 | mfeat-morphological | 2000 | 7 | 10 |
| 4 | anneal_ORIG | 898 | 39 | 6 | 62 | mfeat-pixel | 2000 | 241 | 10 |
| 5 | AR10P_130_674 | 130 | 675 | 10 | 63 | mfeat-zernike | 2000 | 48 | 10 |
| 6 | arrhythmia | 452 | 280 | 16 | 64 | molecular-biology_promoters | 106 | 59 | 2 |
| 7 | audiology | 226 | 70 | 24 | 65 | monks-problems-1_test | 432 | 7 | 2 |
| 8 | autos | 205 | 26 | 7 | 66 | monks-problems-1_train | 124 | 7 | 2 |
| 9 | balance-scale | 625 | 5 | 3 | 67 | monks-problems-2_test | 432 | 7 | 2 |
| 10 | breast-cancer | 286 | 10 | 2 | 68 | monks-problems-2_train | 169 | 7 | 2 |
| 11 | breast-w | 699 | 10 | 2 | 69 | monks-problems-3_test | 432 | 7 | 2 |
| 12 | bridges_version1 | 107 | 13 | 6 | 70 | monks-problems-3_train | 122 | 7 | 2 |
| 13 | bridges_version2 | 107 | 13 | 6 | 71 | mushroom | 8124 | 23 | 2 |
| 14 | car | 1728 | 7 | 4 | 72 | oh0.wc | 1003 | 3183 | 10 |
| 15 | CLL-SUB-111_111_2856 | 111 | 2857 | 3 | 73 | oh10.wc | 1050 | 3239 | 10 |
| 16 | cmc | 1473 | 10 | 3 | 74 | oh15.wc | 913 | 3101 | 10 |
| 17 | colic | 368 | 23 | 2 | 75 | oh5.wc | 918 | 3013 | 10 |
| 18 | colic.ORIG | 368 | 28 | 2 | 76 | pasture | 36 | 23 | 3 |
| 19 | colon | 62 | 2001 | 2 | 77 | pendigits | 10992 | 17 | 10 |
| 20 | credit-a | 690 | 16 | 2 | 78 | PIE10P_210_1520 | 210 | 1521 | 10 |
| 21 | credit-g | 1000 | 21 | 2 | 79 | postoperative-patient-data | 90 | 9 | 3 |
| 22 | cylinder-bands | 540 | 40 | 2 | 80 | primary-tumor | 339 | 18 | 22 |
| 23 | dermatology | 366 | 35 | 6 | 81 | segment | 2310 | 20 | 7 |
| 24 | diabetes | 768 | 9 | 2 | 82 | shuttle-landing-control | 15 | 7 | 2 |
| 25 | ECML90x27679 | 90 | 27680 | 43 | 83 | sick | 3772 | 30 | 2 |
| 26 | ecoli | 336 | 8 | 8 | 84 | SMK-CAN-187_187_1815 | 187 | 1816 | 2 |
| 27 | Embryonaldataset_C | 60 | 7130 | 2 | 85 | solar-flare_1 | 323 | 13 | 2 |
| 28 | eucalyptus | 24863 | 249 | 12 | 86 | solar-flare_2 | 1066 | 13 | 3 |
| 29 | flags | 194 | 30 | 8 | 87 | sonar | 208 | 61 | 2 |
| 30 | GCM_Test | 46 | 16064 | 14 | 88 | soybean | 683 | 36 | 19 |
| 31 | gina_agnostic | 3468 | 971 | 2 | 89 | spectf_test | 269 | 45 | 2 |
| 32 | gina_prior | 3468 | 785 | 2 | 90 | spectf_train | 80 | 45 | 2 |
| 33 | gina_prior2 | 3468 | 785 | 10 | 91 | spectrometer | 531 | 103 | 48 |
| 34 | glass | 214 | 10 | 7 | 92 | spect_test | 187 | 23 | 2 |
| 35 | grub-damage | 155 | 9 | 4 | 93 | spect_train | 80 | 23 | 2 |
| 36 | heart-c | 303 | 14 | 5 | 94 | splice | 3190 | 62 | 3 |
| 37 | heart-h | 294 | 14 | 5 | 95 | sponge | 76 | 46 | 3 |
| 38 | heart-statlog | 270 | 14 | 2 | 96 | squash-stored | 52 | 25 | 3 |
| 39 | hepatitis | 155 | 20 | 2 | 97 | squash-unstored | 52 | 24 | 3 |
| 40 | hypothyroid | 3772 | 30 | 4 | 98 | sylva_agnostic | 14395 | 217 | 2 |
| 41 | ionosphere | 351 | 35 | 2 | 99 | sylva_prior | 14395 | 109 | 2 |
| 42 | iris | 150 | 5 | 3 | 100 | TOX-171_171_1538 | 171 | 1538 | 4 |
| 43 | kdd_ipums_la_97-small | 7019 | 61 | 9 | 101 | tr11.wc | 414 | 6430 | 9 |
| 44 | kdd_ipums_la_98-small | 7485 | 61 | 10 | 102 | tr12.wc | 313 | 5805 | 8 |
| 45 | kdd_ipums_la_99-small | 8844 | 61 | 9 | 103 | tr23.wc | 204 | 5833 | 6 |
| 46 | kdd_JapaneseVowels_test | 5687 | 15 | 9 | 104 | tr31.wc | 927 | 10129 | 7 |
| 47 | kdd_JapaneseVowels_train | 4274 | 15 | 9 | 105 | tr41.wc | 878 | 7455 | 10 |
| 48 | kdd_synthetic_control | 600 | 62 | 6 | 106 | tr45.wc | 690 | 8262 | 10 |
| 49 | kr-vs-kp | 3196 | 37 | 2 | 107 | trains | 10 | 33 | 2 |
| 50 | labor | 57 | 17 | 2 | 108 | vehicle | 846 | 19 | 4 |
| 51 | Leukemia | 72 | 7130 | 2 | 109 | vote | 435 | 17 | 2 |
| 52 | Leukemia_3c | 72 | 7130 | 3 | 110 | vowel | 990 | 14 | 11 |
| 53 | leukemia_test_34x7129 | 34 | 7130 | 2 | 111 | wap.wc | 1560 | 8461 | 20 |
| 54 | leukemia_train_38x7129 | 38 | 7130 | 2 | 112 | waveform-5000 | 5000 | 41 | 3 |
| 55 | lung-cancer | 32 | 57 | 3 | 113 | white-clover | 63 | 32 | 4 |
| 56 | lymph | 148 | 19 | 4 | 114 | wine | 178 | 14 | 3 |
| 57 | Lymphoma45x4026+2classes | 45 | 4027 | 2 | 115 | zoo | 101 | 18 | 7 |
| 58 | Lymphoma96x4026+10classes | 96 | 4027 | 11 | | | | | |

Table 2: Statistical summary of the 115 data sets

## 4.2 Experimental Setup

In order to evaluate the performance of the proposed FSS algorithm recommendation method, further verify whether the proposed method is potentially useful in practice, and confirm the reproducibility of our experiments, we set the experimental study as follows.

### 4.2.1 FSS Algorithms

FSS algorithms can be grouped into two broad categories: Wrapper and Filter (Molina et al., 2002; Kohavi & John, 1997). The Wrapper method uses the error rate of the classification algorithm as the evaluation function to measure a feature subset, while the evaluation function of the Filter method is independent of the classification algorithm. The accuracy of the Wrapper method is usually high; however, the generality of the result is limited, and the computational complexity is high. In comparison, Filter method is of generality, and the computational complexity is low. Due to the fact that the Wrapper method is computationally expensive (Dash & Liu, 1997), the Filter method is usually a good choice when the number of features is very large. Thus, we focus on the Filter method in our experiment.

A number of Filter based FSS algorithms have been proposed to handle feature selection problems. These algorithms can be significantly distinguished by i) the search method used to generate the feature subset being evaluated, and ii) the evaluation measures used to assess the feature subset (Liu & Yu, 2005; de Souza, 2004; Dash & Liu, 1997; Pudil, Novovičová, & Kittler, 1994).

In order to guarantee the generality of our experimental results, twelve well-known or the latest search methods and four representative evaluation measures are employed. The brief introduction of these search methods and evaluation measures is as follows.

1) *Search methods*

   i) Sequential forward search (SFS): Starting from the empty set, sequentially add the feature which results in the highest value of objective function into the current feature subset.

   ii) Sequential backward search (SBS): Starting from the full set, sequentially eliminate the feature which results in smallest or no decrease in the value of objective function from the current feature subset.

   iii) Bi-direction search (BiS): A parallel implementation of SFS and SBS. It searches the feature subset space in two directions.

   iv) Genetic search (GS): A randomized search method which performs using a simple genetic algorithm (Goldberg, 1989). The genetic algorithm finds the feature subset to maximize special output function using techniques inspired by natural evolution.

   v) Linear search (LS): An extension of BestFirst search (Gutlein, Frank, Hall, & Karwath, 2009) which searches the space of feature subsets by greedy hill-climbing augmented with a backtracking facility.

   vi) Rank search (RS) (Battiti, 1994): It uses a feature evaluator (such as gain ratio) to rank all the features. After a feature evaluator is specified, a forward selection search is used to generate a ranking list.

   vii) Scatter search (SS) (Garci'a Lopez, Garci'a Torres, Melian Batista, Moreno Perez, & Moreno-Vega, 2006): This method performs a scatter search through the feature subset space. It starts with a population of many significant and diverse feature subsets, and stops when the assessment criteria is higher than a given threshold or does not have improvement any longer.

   viii) Stepwise search (SWS) (Kohavi & John, 1997): A variation of the forward search. At each step in the search process, after a new feature is added, a test is performed to check if some features can be eliminated without significant reduction in the output function.

   ix) Tabu search (TS) (Hedar, Wang, & Fukushima, 2008): It is proposed for combinatorial optimization problems. It is an adaptive memory and responsive exploration by combining a local search process with anti-cycling memory-based rules to avoid trapping in local optimal solutions.

x) Interactive search (Zhao & Liu, 2007): It traverses the feature subset space for maximizing the target function while taking consideration the interaction among features.

xi) FCBF search (Yu & Liu, 2003): It evaluates features via the relevance and redundancy analysis, and uses the analysis results as guideline to choose features.

xii) Ranker (Kononenko, 1994; Kira & Rendell, 1992; Liu & Setiono, 1995): It evaluates each feature individually and ranks the features by the values of their evaluation metrics.

2) *Evaluation measures*

i) Consistency (Liu & Setiono, 1996; Zhao & Liu, 2007): This kind of measure evaluates the worth of a feature subset by the level of consistency in the target concept when the instances are projected onto the feature subset. The consistency of any feature subset can never be lower than that of the full feature set.

ii) Dependency (Hall, 1999; Yu & Liu, 2003): This kind of measure evaluates the worth of a subset of features by considering the individual predictive ability of each feature along with the degree of redundancy among these features. The FSS methods based on this kind of measure assume that good feature subsets contain features closely correlated with the target concept, but uncorrelated with each other.

iii) Distance (Kira & Rendell, 1992; Kononenko, 1994): This kind of measure is proposed based on the assumption that the distance of instances from different target concepts is greater than that from same target concepts.

iv) Probabilistic significance (Zhou & Dillon, 1988; Liu & Setiono, 1995): This measure evaluates the worth of a feature by calculating the probabilistic significance as a two-way function, i.e., the association between feature and the target concept. A good feature should have significant association with the target concept.

We should pay attention to that, besides the above four evaluation measures, there is another basic kind of measure: information-based measure (Liu & Yu, 2005; de Souza, 2004; Dash & Liu, 1997), which is not contemplated in the experiments. The reason is demonstrated as follow. The information-based measure is usually in conjunction with ranker search method. Thus, the FSS algorithms based on this kind of measure usually provide a rank list of the features instead of telling us which features are relevant to the learning target. In this case, we should preassign particular thresholds for these FSS algorithms to pick up the relevant features. However, there is not any effective method to set the thresholds or any acknowledged default threshold for these FSS algorithms. Moreover, it is unfair to conclude that these information measure based FSS algorithms with any assigned threshold are not appropriate when comparing to the other FSS algorithms. Therefore, this kind of FSS algorithm is not employed in our experiments.

Based on the search methods and the evaluation measures introduced above, 22 different FSS algorithms are obtained. Table 3 shows the brief introduction of these FSS algorithms. Where all these algorithms are available in the data mining toolkit Weka[5] (Hall, Frank,

---

5. http://www.cs.waikato.ac.nz/ml/weka/

Holmes, Pfahringer, Reutemann, & Witten, 2009), and the search method INTERACT is implemented based on Weka and its source codes are available online[6].

| ID | Search Method | Evaluation Measure | Notation | ID | Search Method | Evaluation Measure | Notation |
|----|---------------|--------------------|----------|----|---------------|--------------------|----------|
| 1 | BestFirst + Sequential Forward Search | Dependency | CFS-SFS | 12 | BestFirst + Sequential Backward Search | Consistency | Cons-SBS |
| 2 | BestFirst + Sequential Backward Search | Dependency | CFS-SBS | 13 | BestFirst + Bi-direction Search | Consistency | Cons-BiS |
| 3 | BestFirst + Bi-direction Search | Dependency | CFS-BiS | 14 | Genetic Search | Consistency | Cons-GS |
| 4 | Genetic Search | Dependency | CFS-GS | 15 | Linear Search | Consistency | Cons-LS |
| 5 | Linear Search | Dependency | CFS-LS | 16 | Rank Search | Consistency | Cons-RS |
| 6 | Rank Search | Dependency | CFS-RS | 17 | Scatter Search | Consistency | Cons-SS |
| 7 | Scatter Search | Dependency | CFS-SS | 18 | Stepwise Search | Consistency | Cons-SWS |
| 8 | Stepwise Search | Dependency | CFS-SWS | 19 | Interactive Search | Consistency | INTERACT-C |
| 9 | Tabu Search | Dependency | CFS-TS | 20 | FCBFsearch | Dependency | FCBF |
| 10 | Interactive Search | Dependency | INTERACT-D | 21 | Ranker | Distance | Relief-F |
| 11 | BestFirst + Sequential Forward Search | Consistency | Cons-SFS | 22 | Ranker | Probabilistic Significance | Signific |

Table 3: Introduction of the 22 FSS algorithms

It is noted that some of these algorithms require particular settings of certain parameters. For the purpose of allowing other researchers to confirm our results, we introduce the parameter settings of these FSS algorithms. Such as, for FSS algorithms "INTERACT-D" and "INTERACT-C", there is a parameter, *c-contribution* threshold, used to identify the irrelevant features. We set this threshold as 0.0001 suggested by Zhao and Liu (2007). For FSS algorithm "FCBF", we set the relevance threshold to be the $SU$ (Symmetric Uncertainty) value of the $\lfloor N/\log N \rfloor$th ranked feature suggested by Yu and Liu (2003). For FSS algorithm "Relief-F", we set the significance threshold to 0.01 used by Robnik-Šikonja and Kononenko (2003). For FSS algorithm "Signific", there is a threshold, statistical significance level $\alpha$, used to identify the irrelevant features. We set $\alpha$ as the commonly-used value 0.01 in our experiment. The other FSS algorithms are conducted in the Weka environment with the default setting(s).

### 4.2.2 CLASSIFICATION ALGORITHMS

Since the actual relevant features of real world data sets are usually not known in advance, it is impracticable to directly evaluate an FSS algorithm by the selected features. Classification accuracy is an extensively used metric for evaluating the performance of FSS algorithms, and also plays an important role in our proposed performance metric $EARR$ for assessing different FSS algorithms.

However, different classification algorithms have different biases. An FSS algorithm may be more suitable for some classification algorithms than others (de Souza, 2004). This fact affects the performance evaluation of FSS algorithms.

With this in mind, in order to demonstrate that our proposed FSS algorithm recommendation method is not limited to any particular classification algorithm, five representative classification algorithms based on different hypotheses are employed. They are bayes-based Naive Bayes (John & Langley, 1995) and Bayes Network (Friedman, Geiger, & Goldszmidt, 1997), information gain-based C4.5 (Quinlan, 1993), rule-based PART (Frank & Witten, 1998), and instance-based IB1 (Aha, Kibler, & Albert, 1991), respectively.

Although Naive Bayes and Bayes Net are both bayes-based classification algorithms, they are quite different from each other since Naive Bayes is proposed based on the hypoth-esis that the features are conditional independent (John & Langley, 1995), while Bayes Net takes into account the feature interaction (Friedman et al., 1997).

---

6. http://www.public.asu.edu/huanliu/INTERACT/INTERACTsoftware.html

### 4.2.3 Measures to Evaluate the Recommendation Method

FSS algorithm recommendation is an application of meta-learning. So far as we know, there are no unified measures to evaluate the performance of the meta-learning methods. In order to assess our proposed FSS algorithm recommendation method, two measures, *recommendation hit ratio* and *recommendation performance ratio*, are defined.

Let $D$ be a given data set and $A_{rec}$ be an FSS algorithm recommended by the recommendation method for $D$, these two measures can be introduced as follows.

1) *Recommendation hit ratio*

An intuitive evaluation criterion is whether the recommended FSS algorithm $A_{rec}$ meets users' requirements. That is, whether $A_{rec}$ is the optimal FSS algorithm for $D$, or the performance of $A_{rec}$ on $D$ has no significant difference with that of the optimal FSS algorithm.

Suppose $A_{opt}$ represents the optimal FSS algorithm for $D$, and ASet$_{opt}$ denotes the FSS algorithm set in which each algorithm has no significant difference with $A_{opt}$ (of course it includes $A_{opt}$ as well). Then, a measure named *recommendation hit* can be defined to assess whether the recommended algorithm $A_{rec}$ is effective on $D$.

**Definition 1** *(Recommendation hit). If an FSS algorithm $A_{rec}$ is recommended to a data set $D$, then the recommendation hit $Hit(A_{rec}, D)$ is defined as*

$$Hit(A_{rec}, D) = \begin{cases} 1, if \ A_{rec} \in ASet_{opt} \\ 0, otherwise \end{cases}. \tag{6}$$

Where $Hit(A_{rec}, D) = 1$ means the recommendation is effective since the recommended FSS algorithm $A_{rec}$ is one of the algorithms in ASet$_{opt}$ for $D$, while $hit(A_{rec}, D) = 0$ indicates the recommended FSS algorithm $A_{rec}$ is not a member of ASet$_{opt}$, i.e., $A_{rec}$ is significantly worse than the optimal FSS algorithm $A_{opt}$ on $D$, thus the recommendation is bad.

From Definition 1 we know that the recommendation hit $Hit(A_{rec}, D)$ is used to evaluate the recommendation method for a single data set. Thus, it is extended as recommendation hit ratio to evaluate the recommendation for a set of data sets, and is defined as follows.

**Definition 2** *Recommendation hit ratio*

$$Hit\_Ratio(A_{rec}) = \frac{1}{G} \sum_{i=1}^{G} Hit(A_{rec}, D_i). \tag{7}$$

Where $G$ is the number of the historical data sets, e.g., $G = 115$ in our experiment.

Definition 2 represents the percentage of data sets on which the appropriate FSS algorithms are effectively recommended by our recommendation method. The larger this value, the better the recommendation method.

2) *Recommendation performance ratio*

The recommendation hit ratio reveals that whether or not an appropriate FSS algorithm is recommended for a given data set, but it cannot tell us the margin of the recommended algorithm to the best one. Thus, a new measure, the recommendation performance ratio for a recommendation, is defined.

**Definition 3** (*Recommendation performance ratio*). *Let $EARR_{rec}$ and $EARR_{opt}$ be the performance of the recommended FSS algorithm $A_{rec}$ and the optimal FSS algorithm on $D$, respectively. Then, the recommendation performance ratio (RPR) for $A_{rec}$ is defined as*

$$RPR(A_{rec}, D) = EARR_{rec}/EAAR_{opt}. \tag{8}$$

In this definition, the best performance $EARR_{opt}$ is employed as a benchmark. Without the benchmark, it is hard to determine the recommended algorithms are good or not. For example, suppose the $EARR$ of $A_{rec}$ on $D$ is 0.59. If $EARR_{opt} = 0.61$, then the recommendation is effective since $EARR$ of $A_{rec}$ is very close to $EARR_{opt}$. However, the recommendation is poor if $EARR_{opt} = 0.91$.

$RPR$ is the ratio of $EARR$ of a recommended FSS algorithm to that of the optimal one. It measures how close the recommended FSS algorithm to the optimal one, and reveals the relative performance of the recommended FSS algorithm. Its value varies from 0 to 1, and the larger the value of $RPR$, the closer the performance of the recommended FSS algorithm to that of the optimal one. The recommended algorithm is the optimal one if and only if $RPR = 1$.

### 4.2.4 Values of the Parameters $\alpha$ and $\beta$

In this paper, a multi-criteria metric $EARR$ is proposed to evaluate the performance of an FSS algorithm. For the proposed metric $EARR$, two parameters $\alpha$ and $\beta$ are established for users to express their requirements on algorithm performance.

In the experiment, when presenting the results, two representative value pairs of parameters $\alpha$ and $\beta$ are used as follows:

1) $\alpha = 0$ and $\beta = 0$. This setting represents the situation where the classification accuracy is most important. The higher the classification accuracy over the selected features, the better the corresponding FSS algorithms.

2) $\alpha \neq 0$ and $\beta \neq 0$. This setting represents the situation where the user can tolerate an accuracy attenuation and favor the FSS algorithms with shorter runtime and fewer selected features. In the experiment, both $\alpha$ and $\beta$ are set to 10% that is quite different from $\alpha = \beta = 0$. This allows us can explore the impact of these two parameters on the recommendation results.

Moreover, in order to explore how parameters $\alpha$ and $\beta$ affect the recommended FSS algorithms in terms of classification accuracy, runtime and the number of selected features, different parameters settings are provided. Specifically, the values of $\alpha$ and $\beta$ vary from 0 to 10% with an increase of 1%.

## 4.3 Experimental Process

In order to make sure the soundness of our experimental conclusion and guarantee the experiments reported being reproducible, in this part, we introduce the four crucial processes used in our experiments. They are i) meta-knowledge database construction, ii) optimal FSS algorithm set identification for a given data set, iii) Recommendation method validation and iv) sensitivity analysis of the number of the nearest data sets on recommendations.

1) *Meta-knowledge database construction*

---

**Procedure** `PerformanceEvaluation`

---

**Inputs** : $data$ = a given data set, i.e, one of the 115 data sets;

$learner$ = a given classification algorithm, i.e., one of {Naive Bayes, C4.5, PART, IB1 or Bayes Network};

$FSSAlgSet$ = {$FSSAlg_1$, $FSSAlg_2$, $\cdots$, $FSSAlg_{22}$}, the set of the 22 FSS algorithms;

**Output** : $EARRset$ = {$EARR_1$, $EARR_2$, $\cdots$, $EARR_{22}$}, the $EARR$s of the 22 $FSS$ algorithms on $data$;

**1** $M = 5$; $FOLDS = 10$;

**2 for** $i = 1$ **to** $22$ **do**

**3**    $EARR_i = 0$;

**4 for** $i = 1$ **to** M **do**

**5**    randomized order from $data$;

**6**    generate $FOLDS$ bins from $data$;

**7**    **for** $j = 1$ **to** $FOLDS$ **do**

**8**      $TestData$ = bin[$j$];

**9**      $TrainData$ = data- $TestData$;

**10**      $numberList$ = Null, $runtimeList$ = Null, $accuracyList$ = Null;

**11**      **for** $k = 1$ **to** $22$ **do**

**12**        ($Subset$, $runtime$) = apply $FSSAlg$ $_k$ on $TrainData$;

**13**        $number$ = |$Subset$ |;

**14**        $RedTestData$ = reduce $TestData$ according to selected $Subset$;

**15**        $RedTrainData$ = reduce $TrainData$ according to selected $Subset$;

**16**        $classifier$ = learner ($RedTrainData$);

**17**        $accuracy$ = apply $classifier$ to $RedTestData$;

**18**        $numberList$ [$k$] = $number$, $runtimeList$ [$k$] = $runtime$, $accuracyList$ [$k$] = $accuracy$;

**19**      **for** $k = 1$ **to** $22$ **do**

**20**        $EARR$ = EARRCompution($accuracyList$, $runtimeList$, $numberList$, $k$);

         *//Compute EARR of FSSAlg$_k$ on jth bin of pass i according Eqs. (1) and (2)*

**21**        $EARR_k = EARR_k + EARR$;

**22 for** $i \leftarrow 1$ **to** $22$ **do**

**23**    $EARR_i = EARR_i/(M \times FOLDS)$;

**24 return** $EARRset$;

---

For each data set $D_i$ ($1 \leq i \leq 115$), we i) extract its meta-features $F_i$; ii) calculate the $EARR$s for the 22 candidate FSS algorithms with the *stratified 5× 10-fold cross-validation* strategy (Kohavi, 1995), and iii) combine the meta-features $F_i$ and the $EARR$ of each FSS algorithm together to form a tuple, which is finally added to the meta-knowledge database.

Since the extraction of meta-features and the combination of the meta-features and the $EARR$s are straightforward, we just present the calculation of $EARR$s, procedure *PerformanceEvaluation* shows the details.

2) *Optimal FSS algorithm set identification*

The optimal FSS algorithm set for a given data set $D_i$ consists of the optimal FSS algorithm for this data set and those algorithms that have no significant performance difference with the optimal one on $D_i$.

The optimal FSS algorithm set for a given data set $D_i$ is obtained via a non-parametric Friedman test (1937) followed by a Holm procedure test (1988) on the performance, which is estimated by the 5×10 cross validation strategy, of the 22 FSS algorithms. If the result of the Friedman test shows that there is no significant performance difference among the 22 FSS algorithms, these 22 FSS algorithms are added to the optimal FSS algorithm set. Otherwise, the FSS algorithm with the highest performance is viewed as the optimal one and added to the optimal FSS algorithm set. Then, the Holm procedure test is performed to identify the algorithms from the rest 21 FSS algorithms. The algorithms that have no significant performance differences with the optimal one are added into the optimal FSS algorithm set.

The reason why the non-parametric test is employed lies in that it is difficult for the performance values to follow the normal distribution and satisfy variance homogeneous condition.

Note that the optimal FSS algorithm sets for different settings of parameters $\alpha$ and $\beta$ are different, since the values of these two parameters directly affect the required performance values.

3) *Recommendation method validation*

The *leave-one-out* strategy is used to empirically evaluate our proposed FSS algorithm recommendation method as follows: for each data set $D_i$ ($1 \leq i \leq 115$) that is viewed as the *test data*, i) identify its $k$ nearest data sets from the *training data* = $\{D_1, \cdots, D_{i-1}, D_{i+1}, \cdots, D_{115}\}$ excluding $D_i$; ii) calculate the performance of the 22 candidate FSS algorithms according to Eq. (5) based on the $k$ nearest data sets where the value of $k$ is determined by the standard cross-validation strategy, and recommend the top three to $D_i$; and iii) evaluate the recommendations by the measures introduced in section 4.2.3.

4) *Sensitivity analysis of the number of the nearest data sets on recommendations*

In order to explore the effect of the number of the nearest data sets on the recommendations and provide users an empirical method to choose its value, for each data set, all the possible numbers of the nearest data sets are tested. That is, when identifying the $k$ nearest data sets for a given data set, $k$ is set from 1 to the number of the historical data sets minus 1 (e.g., 114 in this experiment).

## 4.4 Results and Analysis

In this section, we present the recommendation results in terms of recommended FSS algorithms, hit ratio and performance ratio , respectively. Due to the space limit of the paper, we do not list all the recommendations, but instead present the results under two significantly different pairs of $\alpha$ and $\beta$, i.e., ($\alpha = 0$, $\beta = 0$) and ($\alpha = 10\%$, $\beta = 10\%$).

Afterward, we also provide the experimental results of the influence of the user-oriented parameters $\alpha$ and $\beta$ on recommendations in terms of classification accuracy, runtime, and the number of selected features, respectively.

### 4.4.1 Recommended Algorithms and Hit Ratio

Figs. 2, 3, 4, 5 and 6 show the first recommended FSS algorithms for the 115 data sets when the classification algorithms Naive Bayes, C4.5, PART, IB1 and Bayes Network are used, respectively.

In each figure, there are two sub-figures corresponding to the recommendation results for $(\alpha = 0, \beta = 0)$ and $(\alpha = 10\%, \beta = 10\%)$, respectively. In each sub-figure, '○' and '×' denote the correctly and incorrectly recommended algorithms, respectively.

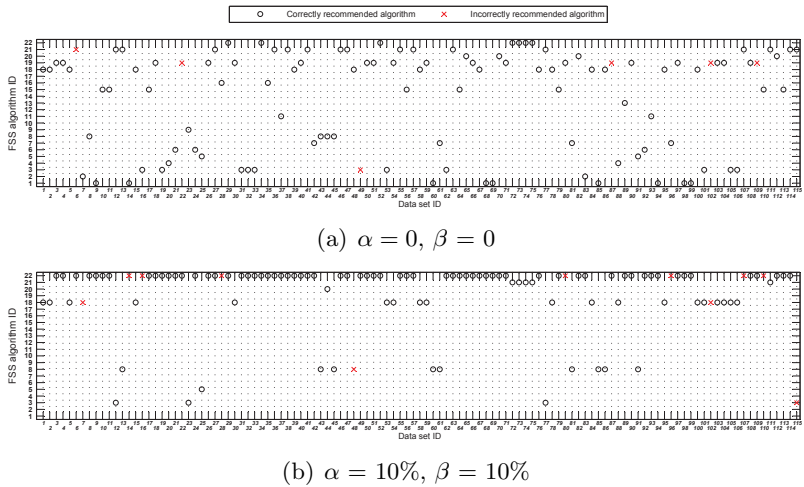

(a) $\alpha = 0$, $\beta = 0$

(b) $\alpha = 10\%$, $\beta = 10\%$

Figure 2: FSS algorithms recommended for the 115 data sets when Naive Bayes is used

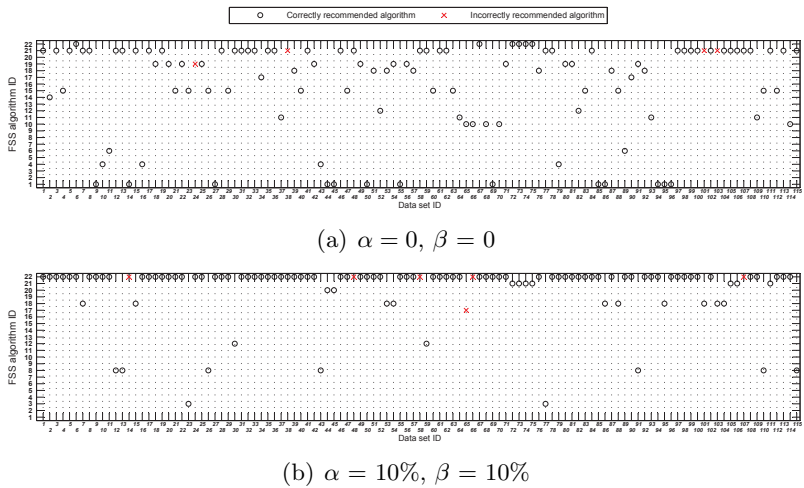

(a) $\alpha = 0$, $\beta = 0$

(b) $\alpha = 10\%$, $\beta = 10\%$

Figure 3: FSS algorithms recommended for the 115 data sets when C4.5 is used

From these figures, we observe that:

1) For all the five classifiers, the proposed method can effectively recommend appropriate FSS algorithms for most of the 115 data sets.

In the case of $(\alpha = 0, \beta = 0)$, the number of data sets, whose appropriate FSS algorithms are correctly recommended, is 109 out of 115 for Naive Bayes, 111 out of 115 for C4.5, 109 out of 115 for PART, 108 out of 115 for IB1, and 109 out of 115 for Bayes

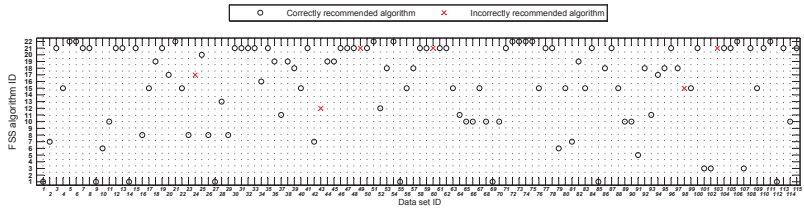

(a) $\alpha = 0,\ \beta = 0$

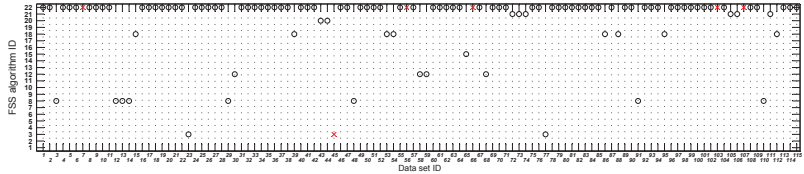

(b) $\alpha = 10\%,\ \beta = 10\%$

Figure 4: FSS algorithms recommended for the 115 data sets when PART is used

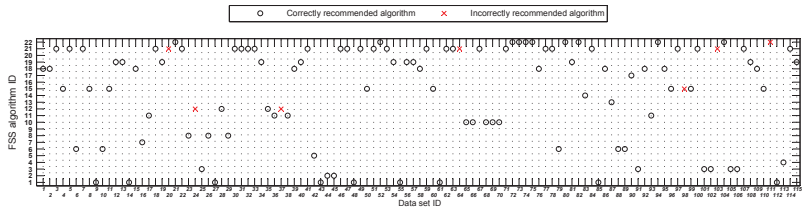

(a) $\alpha = 0,\ \beta = 0$

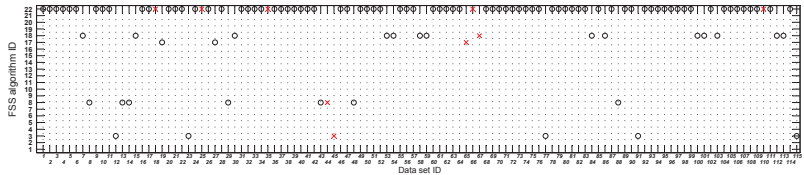

(b) $\alpha = 10\%,\ \beta = 10\%$

Figure 5: FSS algorithms recommended for the 115 data sets when IB1 is used

Network, respectively. This states that the recommendation method is effective when classification accuracy is most important.

In the case of ($\alpha = 10\%,\ \beta = 10\%$), the number of data sets, whose appropriate FSS algorithms are correctly recommended, is 104 out of 115 for Naive Bayes, 109 out of 115 for C4.5, 110 out of 115 for PART, 106 out of 115 for IB1, and 104 out of 115 for Bayes Network, respectively. This indicates that the recommendation method also works well when tradeoff is required among classification accuracy, runtime, and the number of selected features.

2) The distribution of the recommended FSS algorithms for the 115 data sets is different for different parameters settings. The distribution is relatively uniform for ($\alpha = 0, \beta = 0$), while it is seriously biased to some algorithm (e.g., the 22th FSS algorithm) for ($\alpha = 10\%$, $\beta = 10\%$).

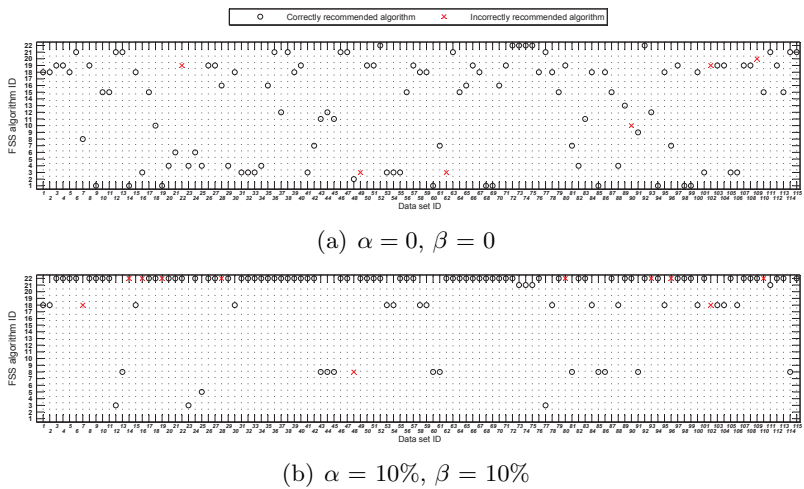

(a) $\alpha = 0$, $\beta = 0$

(b) $\alpha = 10\%$, $\beta = 10\%$

Figure 6: FSS algorithms recommended for the 115 data sets when Bayes Network is used

This phenomenon is similar for all the five classifiers. This can be explained as follows. The FSS algorithms with the best classification accuracy distribute on the 115 data sets uniformly. Thus, in the case of ($\alpha = 0, \beta = 0$) where users favor accurate classifiers, the distribution of the recommended FSS algorithms is relatively uniform as well. However, there exist some FSS algorithms that run faster (e.g., the 22th algorithm "Signific") or select fewer features (e.g., the 8th algorithm "CFS-SWS", the 18th algorithm "Cons-SWS", and the 22th algorithm "Signific") on most of the 115 data sets. For this reason, in the case of ($\alpha = 10\%$, $\beta = 10\%$) where users prefer the FSS algorithms with less runtime and fewer features, the distribution of the FSS algorithms with the best performance on the 115 data sets is biased to some algorithms, so is the recommended FSS algorithms.

3) The 22th FSS algorithm performs well on about 85 out of 115 data sets for all classifiers when ($\alpha = 10\%$, $\beta = 10\%$). It seems that this FSS algorithm is a generally well-performed FSS algorithm that can be adopted by all FSS tasks and there is no need for FSS algorithm recommendation. Unfortunately, this is not the case. The 22th FSS algorithm is still failing to perform well over about a quarter of the 115 data sets. Yet, our recommendation method can distinguish these data sets and further effectively recommend appropriate FSS algorithms for them. This indicates our recommendation method is still necessary in this case.

Compared with ($\alpha = 0$, $\beta = 0$), we can know that this case is due to the larger $\alpha$ and $\beta$ values and can be explained as follows. For all the 22 FSS algorithms, although the classification accuracies of a classifier over the features selected by them are different, the differences are usually bounded. Meanwhile, from Eq. (1) we know that when $\alpha/\beta$ is set to be greater than the bound value, the value of $EARR$ will be dominated by the runtime/number of selected features. This means that if $\alpha$ or $\beta$ is set to be a relatively large value, the algorithm with a lower time complexity or the algorithm that chooses smaller number of features will be recommended, and the classification accuracy over the selected features will be ignored. However, as we know, one of the most important targets of feature selection is to improve the performance of learning algorithms. So it is unreasonable to ignore the classification accuracy and just focus on the speed and the simplicity of an FSS algorithm.

Thus, in real applications, the values of $\alpha$ and $\beta$ should be set under the limit of classification accuracies. Generally, the $\alpha/\beta$ should be bounded by $(acc_{max} - acc_{min})/acc_{min}$, where $acc_{max}$ and $acc_{min}$ denote the maximum and the minimum classification accuracies, respectively.

| Parameter setting | Recommendation | Naive Bayes | C4.5 | PART | IB1 | Bayes Network |
|---|---|---|---|---|---|---|
| | $\text{Alg}_{1st}$ | 94.78 | 96.52 | 94.78 | 93.91 | 94.78 |
| $\alpha = 0, \beta = 0$ | $\text{Alg}_{2nd}$ | 83.48 | 79.13 | 92.17 | 77.39 | 83.48 |
| | $\text{Alg}_{3rd}$ | 74.78 | 80.87 | 84.35 | 75.65 | 73.91 |
| | Top 3 | 99.13 | 98.26 | 99.13 | 99.13 | 98.26 |
| | $\text{Alg}_{1st}$ | 90.43 | 94.78 | 94.78 | 92.17 | 90.43 |
| $\alpha = 10\%, \beta = 10\%$ | $\text{Alg}_{2nd}$ | 71.30 | 69.57 | 70.43 | 64.35 | 71.30 |
| | $\text{Alg}_{3rd}$ | 38.26 | 45.22 | 42.61 | 43.48 | 36.52 |
| | Top 3 | 99.13 | 100.0 | 100.0 | 100.0 | 99.13 |

* $\text{Alg}_x$ denotes only the $x$-th algorithm in the ranking list is recommended while Top 3 means the top three algorithms are recommended.

Table 4: Hit ratio (%) of the recommendations for the five classifiers under different settings of $(\alpha, \beta)$

Table 4 shows the hit ratio of the recommended FSS algorithms for the five classifiers. From it we can observe that:

1) If a single FSS algorithm is recommended, the hit ratio of the first recommended algorithm $\text{Alg}_{1st}$ is the highest, its value is up to 96.52% and at least is 90.43% for all the five classifiers. Thus, $\text{Alg}_{1st}$ should be the first choice.

2) If the top three algorithms are recommended, the hit ratio is up to 100% and at least is 98.62%. That indicates that the confidence of the top three algorithms including an appropriate one is very high. This is the reason why only the top three algorithms are recommended. Moreover, the proposed recommendation method has reduced the number of candidate algorithms to three, users can further pick up the one that fits his/her specific requirement from them.

### 4.4.2 Recommendation Performance Ratio

Figs. 7 and 8 show the recommendation performance ratio $RPR$ of the first recommended FSS algorithm for the five classifiers with $(\alpha = 0, \beta = 0)$ and $(\alpha = 10\%, \beta = 10\%)$, respectively. From these two figures we can observe that, for most data sets and the two settings of $\alpha$ and $\beta$, the $RPRs$ of the recommended FSS algorithms are greater than 95% and some of them are up to 100% no matter which classifier is used. This indicates that the FSS algorithms recommended by our proposed method are very close to the optimal one.

Table 5 shows the average $RPRs$ over the 115 data sets for the five classifiers under different settings of $(\alpha, \beta)$. In this table, for each classifier, columns "Rec" and "Def" shows the $RPR$ value corresponding to the recommended FSS algorithms and default FSS algorithms, respectively. Where the default FSS algorithm is the most frequent best one on the 115 data sets under the classifier.

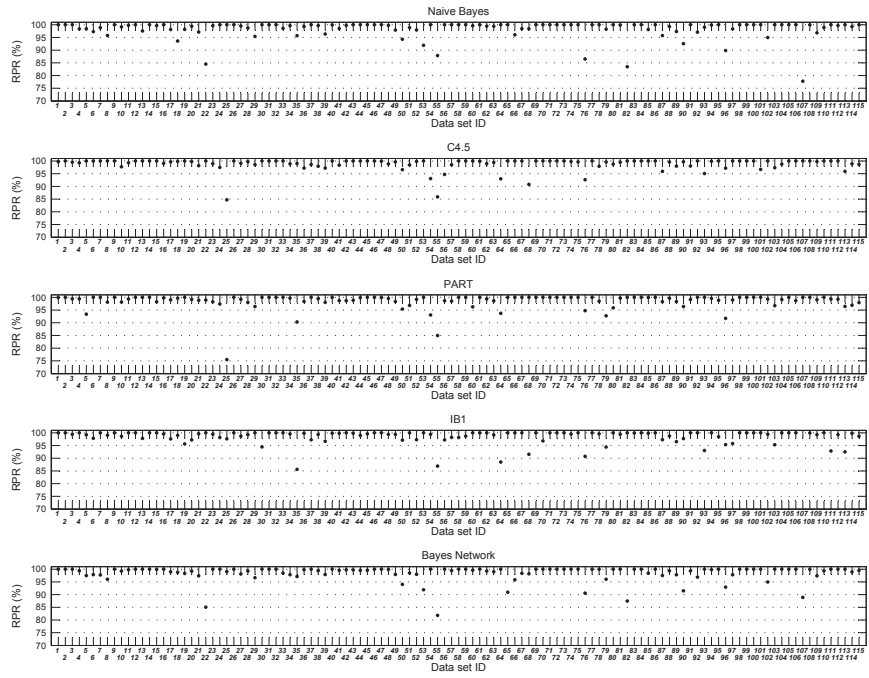

Figure 7: *RPR* of the *1*st recommended FSS algorithm with ($\alpha = 0, \beta = 0$) for the five classifiers

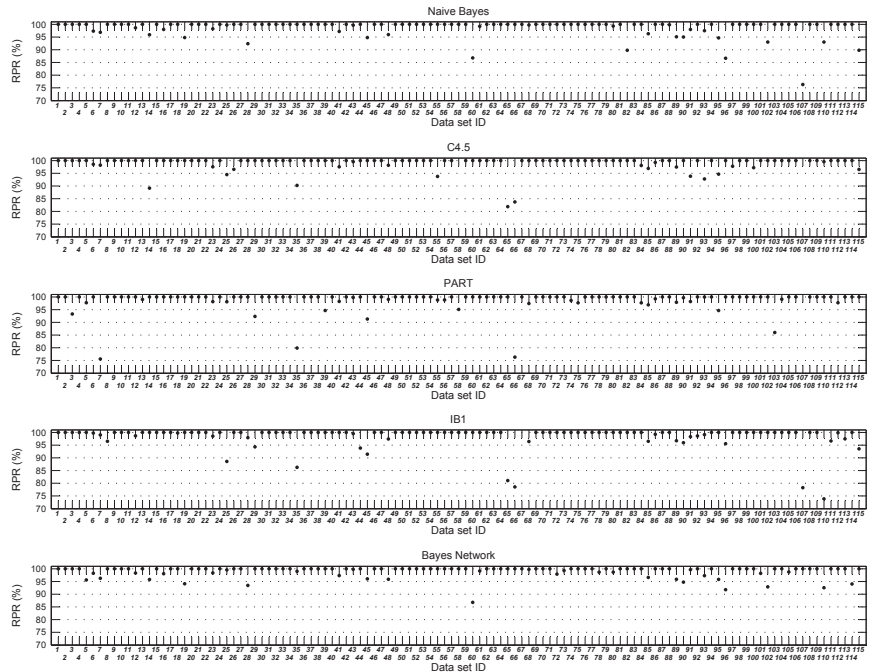

Figure 8: *RPR* of the *1*st recommended FSS algorithm with ($\alpha = 10\%$, $\beta = 10\%$) for the five classifiers

From it we observe that the average $RPR$s range from 97.32% to 98.8% for ($\alpha = 0$, $\beta = 0$), and from 97.82% to 98.99% for ($\alpha = 10\%$, $\beta = 10\%$), respectively. Moreover, the average $RPR$ of the recommended FSS algorithms surpasses that of the default FSS algorithms for all the five different classifiers. This means our proposed recommendation method works very well and greatly fits user's performance requirement.

| Parameter setting | Naive bayes | | C4.5 | | PART | | IB1 | | Bayes Network | |
|---|---|---|---|---|---|---|---|---|---|---|
| | Rec | Def | Rec | Def | Rec | Def | Rec | Def | Rec | Def |
| $\alpha = 0, \beta = 0$ | 98.24 | 96.42 | 98.80 | 98.18 | 97.61 | 94.79 | 97.32 | 96.43 | 98.37 | 96.63 |
| $\alpha = 10\%, \beta = 10\%$ | 98.69 | 91.95 | 97.82 | 92.40 | 97.89 | 92.40 | 98.11 | 92.35 | 98.99 | 92.43 |

Table 5: Average $RPR$ (%) over 115 data sets for the five classifiers

| Data ID | NB | C4.5 | PART | IB1 | BNet | Data ID | NB | C4.5 | PART | IB1 | BNet |
|---|---|---|---|---|---|---|---|---|---|---|---|
| 1 | 0.0443 | 0.0425 | 0.0499 | 0.0423 | 0.0464 | 59 | 0.0446 | 0.0439 | 0.0443 | 0.0477 | 0.0427 |
| 2 | 0.0227 | 0.0131 | 0.0147 | 0.0137 | 0.0129 | 60 | 0.0361 | 0.0351 | 0.0407 | 0.034 | 0.0354 |
| 3 | 0.0118 | 0.0124 | 0.0123 | 0.0117 | 0.0116 | 61 | 0.0087 | 0.0096 | 0.0073 | 0.0078 | 0.0085 |
| 4 | 0.0079 | 0.0092 | 0.0082 | 0.0114 | 0.0096 | 62 | 0.0845 | 0.0843 | 0.0835 | 0.0853 | 0.0859 |
| 5 | 0.0244 | 0.0253 | 0.0257 | 0.0246 | 0.0241 | 63 | 0.0225 | 0.0177 | 0.0211 | 0.02 | 0.0194 |
| 6 | 0.019 | 0.019 | 0.0221 | 0.0192 | 0.0208 | 64 | 0.0108 | 0.0077 | 0.0065 | 0.0076 | 0.0098 |
| 7 | 0.0091 | 0.0093 | 0.0093 | 0.0114 | 0.0113 | 65 | 0.0117 | 0.0105 | 0.0074 | 0.0068 | 0.0071 |
| 8 | 0.0111 | 0.0062 | 0.0076 | 0.0066 | 0.0088 | 66 | 0.0089 | 0.0075 | 0.0058 | 0.008 | 0.0079 |
| 9 | 0.011 | 0.0076 | 0.0072 | 0.0074 | 0.0076 | 67 | 0.0092 | 0.0067 | 0.0079 | 0.0065 | 0.0064 |
| 10 | 0.0091 | 0.0087 | 0.0084 | 0.0138 | 0.0073 | 68 | 0.0082 | 0.0063 | 0.0071 | 0.0085 | 0.0116 |
| 11 | 0.0086 | 0.007 | 0.0072 | 0.0075 | 0.0083 | 69 | 0.0061 | 0.0058 | 0.006 | 0.0067 | 0.0067 |
| 12 | 0.0062 | 0.0068 | 0.0065 | 0.0063 | 0.011 | 70 | 0.0069 | 0.0077 | 0.0079 | 0.0097 | 0.007 |
| 13 | 0.0068 | 0.0085 | 0.0064 | 0.0093 | 0.0096 | 71 | 0.0611 | 0.0496 | 0.051 | 0.0485 | 0.0514 |
| 14 | 0.0077 | 0.0087 | 0.0086 | 0.0084 | 0.0077 | 72 | 0.203 | 0.1993 | 0.1976 | 0.1994 | 0.1983 |
| 15 | 0.0616 | 0.0582 | 0.0575 | 0.058 | 0.0586 | 73 | 0.1854 | 0.1689 | 0.1631 | 0.1652 | 0.1638 |
| 16 | 0.0099 | 0.0083 | 0.0081 | 0.0082 | 0.009 | 74 | 0.1195 | 0.1112 | 0.1105 | 0.1103 | 0.1096 |
| 17 | 0.0074 | 0.0126 | 0.0076 | 0.0084 | 0.0085 | 75 | 0.1246 | 0.1208 | 0.1217 | 0.1209 | 0.1226 |
| 18 | 0.0083 | 0.0077 | 0.0128 | 0.0117 | 0.007 | 76 | 0.0147 | 0.0064 | 0.0068 | 0.0067 | 0.006 |
| 19 | 0.0102 | 0.0078 | 0.0105 | 0.0067 | 0.0091 | 77 | 0.0576 | 0.0526 | 0.0509 | 0.0519 | 0.0533 |
| 20 | 0.007 | 0.0071 | 0.007 | 0.0069 | 0.0081 | 78 | 0.0685 | 0.0704 | 0.0641 | 0.066 | 0.0646 |
| 21 | 0.008 | 0.01 | 0.009 | 0.0075 | 0.0079 | 79 | 0.0081 | 0.006 | 0.0082 | 0.008 | 0.013 |
| 22 | 0.0103 | 0.0079 | 0.011 | 0.0135 | 0.0088 | 80 | 0.0086 | 0.0085 | 0.0087 | 0.0072 | 0.0089 |
| 23 | 0.008 | 0.0083 | 0.0079 | 0.0093 | 0.0072 | 81 | 0.0203 | 0.0144 | 0.0141 | 0.0118 | 0.0184 |
| 24 | 0.0066 | 0.0064 | 0.0067 | 0.0069 | 0.0093 | 82 | 0.0095 | 0.0109 | 0.0054 | 0.0091 | 0.0082 |
| 25 | 0.0088 | 0.0158 | 0.0101 | 0.0082 | 0.0094 | 83 | 0.0244 | 0.0244 | 0.0276 | 0.0257 | 0.0248 |
| 26 | 0.0061 | 0.0068 | 0.0075 | 0.0083 | 0.0077 | 84 | 0.0683 | 0.0671 | 0.0674 | 0.0692 | 0.0664 |
| 27 | 0.0097 | 0.0078 | 0.0113 | 0.0117 | 0.0123 | 85 | 0.0074 | 0.0069 | 0.0086 | 0.0086 | 0.0071 |
| 28 | 0.0083 | 0.0069 | 0.008 | 0.0082 | 0.0074 | 86 | 0.0084 | 0.0077 | 0.0411 | 0.0074 | 0.0132 |
| 29 | 0.007 | 0.0132 | 0.0093 | 0.0072 | 0.007 | 87 | 0.0066 | 0.0101 | 0.0071 | 0.0069 | 0.0077 |
| 30 | 0.0273 | 0.0272 | 0.0291 | 0.0273 | 0.0272 | 88 | 0.0096 | 0.0091 | 0.0124 | 0.014 | 0.0113 |
| 31 | 0.2236 | 0.2231 | 0.2236 | 0.2239 | 0.2255 | 89 | 0.0095 | 0.0095 | 0.0078 | 0.0124 | 0.0092 |
| 32 | 0.2602 | 0.2616 | 0.2605 | 0.262 | 0.2596 | 90 | 0.006 | 0.0069 | 0.0085 | 0.0066 | 0.0079 |
| 33 | 0.3691 | 0.3722 | 0.3689 | 0.3689 | 0.3714 | 91 | 0.0158 | 0.0152 | 0.0164 | 0.0183 | 0.0165 |
| 34 | 0.008 | 0.0103 | 0.0083 | 0.0113 | 0.007 | 92 | 0.0068 | 0.0073 | 0.0066 | 0.0081 | 0.0061 |
| 35 | 0.0084 | 0.0068 | 0.0065 | 0.0064 | 0.009 | 93 | 0.0097 | 0.0076 | 0.008 | 0.0065 | 0.008 |
| 36 | 0.0103 | 0.0068 | 0.0066 | 0.0066 | 0.0069 | 94 | 0.0365 | 0.0396 | 0.0378 | 0.0374 | 0.0375 |
| 37 | 0.0065 | 0.0088 | 0.0086 | 0.0059 | 0.0061 | 95 | 0.0108 | 0.0081 | 0.0064 | 0.0094 | 0.0077 |
| 38 | 0.0084 | 0.007 | 0.0084 | 0.0093 | 0.0085 | 96 | 0.0058 | 0.0062 | 0.0062 | 0.0061 | 0.0064 |
| 39 | 0.0098 | 0.0352 | 0.0069 | 0.0067 | 0.0077 | 97 | 0.0082 | 0.0078 | 0.0064 | 0.0068 | 0.0083 |
| 40 | 0.0278 | 0.0243 | 0.0245 | 0.0246 | 0.0249 | 98 | 0.4402 | 0.4429 | 0.444 | 0.4401 | 0.4413 |
| 41 | 0.007 | 0.0099 | 0.011 | 0.007 | 0.0086 | 99 | 0.4591 | 0.4532 | 0.4557 | 0.4551 | 0.4545 |
| 42 | 0.0138 | 0.006 | 0.0106 | 0.0116 | 0.0063 | 100 | 0.0504 | 0.0496 | 0.0502 | 0.0539 | 0.0526 |
| 43 | 0.1219 | 0.1228 | 0.1214 | 0.1231 | 0.1241 | 101 | 0.1012 | 0.095 | 0.0954 | 0.0966 | 0.0968 |
| 44 | 0.1453 | 0.1427 | 0.144 | 0.145 | 0.1434 | 102 | 0.014 | 0.0393 | 0.0424 | 0.0416 | 0.0394 |
| 45 | 0.1937 | 0.1955 | 0.1972 | 0.194 | 0.1935 | 103 | 0.0633 | 0.0618 | 0.0644 | 0.0635 | 0.0625 |
| 46 | 0.0225 | 0.0232 | 0.0242 | 0.0219 | 0.0228 | 104 | 0.9103 | 0.9096 | 0.9091 | 0.9142 | 0.9122 |
| 47 | 0.0149 | 0.0142 | 0.0125 | 0.0139 | 0.015 | 105 | 0.6484 | 0.6448 | 0.6443 | 0.6464 | 0.6461 |
| 48 | 0.0101 | 0.0125 | 0.0101 | 0.0092 | 0.0104 | 106 | 0.5864 | 0.5884 | 0.5861 | 0.5851 | 0.5854 |
| 49 | 0.0228 | 0.0245 | 0.0191 | 0.0198 | 0.024 | 107 | 0.0067 | 0.0056 | 0.0065 | 0.0075 | 0.0055 |
| 50 | 0.0069 | 0.0075 | 0.0084 | 0.0068 | 0.009 | 108 | 0.0091 | 0.0075 | 0.0102 | 0.0131 | 0.0103 |
| 51 | 0.0195 | 0.0201 | 0.0196 | 0.0209 | 0.0197 | 109 | 0.0082 | 0.0074 | 0.0087 | 0.0064 | 0.0095 |
| 52 | 0.0202 | 0.0207 | 0.0165 | 0.0192 | 0.0165 | 110 | 0.0088 | 0.0131 | 0.0131 | 0.0093 | 0.0098 |
| 53 | 0.0128 | 0.0135 | 0.0116 | 0.0156 | 0.0158 | 111 | 0.7746 | 0.7759 | 0.7736 | 0.7739 | 0.7746 |
| 54 | 0.0128 | 0.013 | 0.0197 | 0.0138 | 0.0134 | 112 | 0.0267 | 0.0255 | 0.025 | 0.0266 | 0.0282 |
| 55 | 0.0075 | 0.0073 | 0.0069 | 0.0065 | 0.0068 | 113 | 0.0082 | 0.0075 | 0.0066 | 0.0063 | 0.0135 |
| 56 | 0.0084 | 0.0073 | 0.007 | 0.006 | 0.0069 | 114 | 0.0106 | 0.0095 | 0.0095 | 0.0055 | 0.0073 |
| 57 | 0.0146 | 0.008 | 0.0095 | 0.0082 | 0.0103 | 115 | 0.0086 | 0.0079 | 0.0073 | 0.0062 | 0.0064 |
| 58 | 0.0432 | 0.0464 | 0.0449 | 0.0436 | 0.0445 | Average | 0.0658 | 0.0651 | 0.0652 | 0.0649 | 0.0650 |

* "NB" and "BNet" denote Naive Bayes and Bayes Network, respectively.

Table 6: Recommendation time over 115 data sets for the five classifiers (in *second*)

### 4.4.3 RECOMMENDATION TIME

When recommending FSS algorithms for a feature selection problem, the recommendation time is contributed by meta-features extraction, $k$ nearest data sets identification, and the candidate algorithm ranking according to their performance on the $k$ data sets.

Of these three recommendation time contributors, only the candidate algorithm ranking is related with the parameters $\alpha$ and $\beta$ of the performance metric $EARR$.

However, the computation of performance $EARR$ is the same whatever the values of $\alpha$ and $\beta$ are. This means recommendation time is independent of the specific settings of $\alpha$ and $\beta$. Thus, we just present the recommendation time with $(\alpha = 0, \beta = 0)$, and Table 6 shows the details.

From Table 6 we observe that for a given data set, the recommendation time differences for the five classifiers are small. The reason is that the recommendation time is mainly contributed by the extraction of meta-features, which has no relation with classifiers. This is consistent with the time complexity analysis in Section 3.2. We also observe that for most of the data sets, the recommendation time is less than 0.1 second, and its average value on the 115 data sets is around 0.65 second for each of the five classifiers. This is much faster than the conventional cross validation method.

### 4.4.4 IMPACT OF THE PARAMETERS $\alpha$ AND $\beta$

Figs. 9, 10, 11, 12 and 13 show the impact of the settings of $\alpha$ and $\beta$ on the classification accuracy, the runtime of feature selection, the number of selected features, the Hit Ratio and the RPR value, respectively.

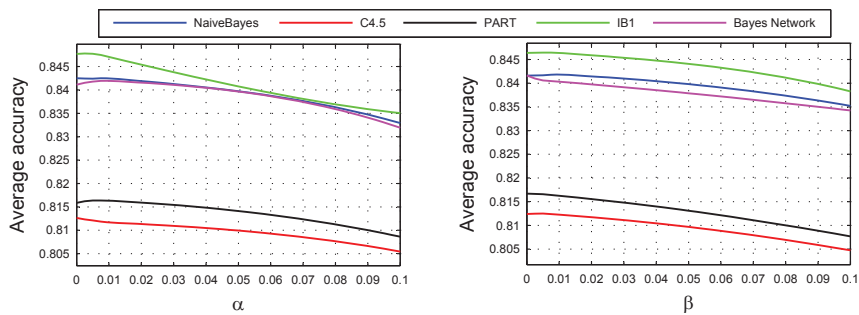

Figure 9: Classification accuracies of the five classifiers with the recommended FSS algorithms under different values of $\alpha$ and $\beta$

Fig. 9 shows the classification accuracies of the five classifiers under the different values of $\alpha$ and $\beta$. From it we observe that, with the increase of either $\alpha$ or $\beta$, the classification accuracies of the five classifiers with the recommended FSS algorithms decrease. This is because the increase of $\alpha$ or $\beta$ indicates that users much prefer faster FSS algorithms or the FSS algorithms that can get less features. Thus, the proportion of classification accuracy in performance is decreased. This means the ranks of the FSS algorithms that run faster and/or get less features are improved and the corresponding FSS algorithms are finally selected.

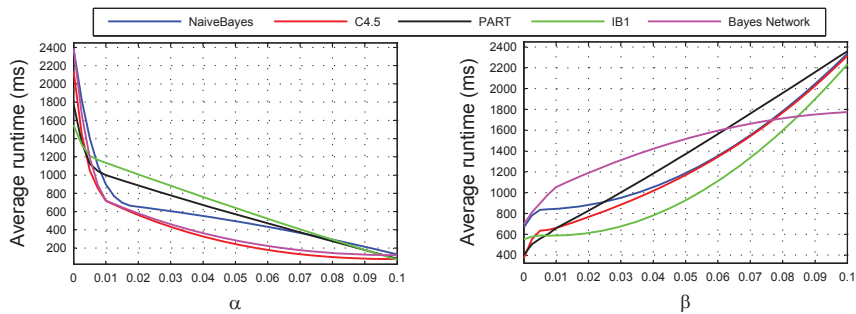

Figure 10: Runtime of the FSS algorithms recommended to the five classifiers under different values of $\alpha$ and $\beta$

Fig. 10 shows the runtime of the FSS algorithms that recommended to the five classifiers under the different values of $\alpha$ and $\beta$ for the five classifiers. From it we observe that:

1) With the increase of $\alpha$, the average runtime of the recommended FSS algorithms for each classifier decreases. Note a larger value of $\alpha$ means users favor faster FSS algorithms. Thus, this indicates that user's performance requirement is met since faster FSS algorithms were recommended.

2) With the increase of $\beta$, the average runtime of the recommended FSS algorithms increases as well. This is because in our proposed recommendation method, the appropriate FSS algorithms for a given data set are recommended based on its nearest data sets. Moreover, in the experiment, for more than half (i.e., 69) of the 115 data sets, there is a negative correlation between the number of selected features and the runtime of the 22 FSS algorithms. Thus, the more data sets with this kind of negative correlation, the more possible the nearest neighbors of a given data set have the negative correlation. Therefore, a larger $\beta$ means longer runtime. Another possible reason is that a larger value of $\beta$ means users favor the FSS algorithms that choose fewer features, and in order to get fewer features, the FSS algorithms need to consume relatively more time.

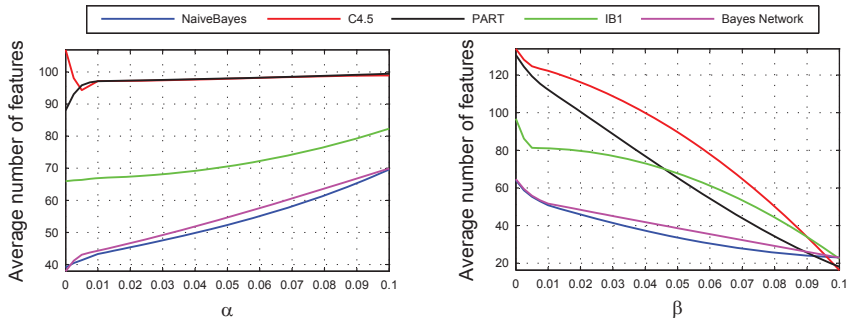

Figure 11: Number of features selected by the FSS algorithms that recommended to the five classifiers under different values of $\alpha$ and $\beta$

Fig. 11 shows the number of features selected by the FSS algorithms that recommended to the five classifiers under different values of $\alpha$ and $\beta$. From it we observe that:

1) With the increase of $\alpha$, the average number of selected features increases as well. This is because in our proposed recommendation method, the appropriate FSS algorithms for a given data set are recommended based on its nearest data sets. Moreover, in the experiment, for more than half (i.e., 69) of the 115 data sets, there is a negative correlation between the number of selected features and the runtime of the 22 FSS algorithms. Thus, the more data sets with this kind of negative correlation, the more possible the nearest neighbors of a given data set have the negative correlation. Therefore, a larger $\alpha$ means more features. Another possible reason is that a larger value of $\alpha$ means users favor faster FSS algorithms. It is possible that shorter computation time can be obtained via filter out less features so more features are remained.

   Note that there is an exception. That is, the average number of selected features for C4.5 decreases when the value of $\alpha$ is small. However, the decrement comes up in a quite small range of $\alpha$ (i.e., $< 0.005$).

2) With the increase of $\beta$, the average number of features selected by the recommended FSS algorithm decreases. Note a larger value of $\beta$ means users favor the FSS algorithms that can get fewer features. Thus, this indicates that user's requirement is met since the FSS algorithms that can get fewer features were recommended.

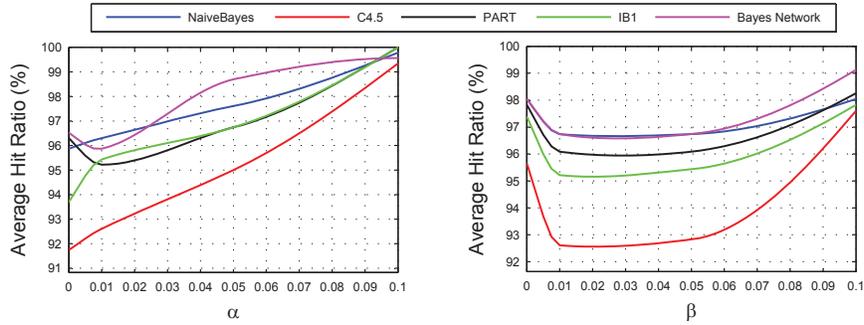

Figure 12: Average Hit Ratio of the FSS algorithms that recommended to the five classifiers under different values of $\alpha$ and $\beta$

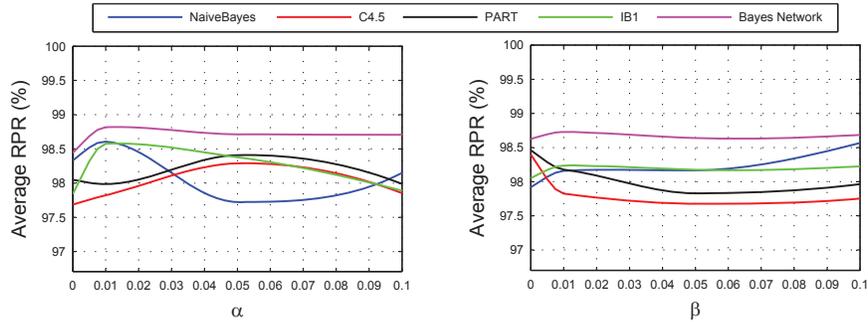

Figure 13: Average RPR of the FSS algorithms that recommended to the five classifiers under different values of $\alpha$ and $\beta$

Figs. 12 and 13 show the average hit ratio and RPR of the recommended FSS algorithms under different values of $\alpha$ and $\beta$ for the five classifiers.

From them we observe that, the average hit ratio falls in the intervals [91.74%, 100%] under $\alpha$ and [92.56%, 99.13%]) under $\beta$. The average RPR varies in the intervals [97.69%, 98.82%] under $\alpha$ and [97.68%, 98.73%] under $\beta$. With the change of the $\alpha$ and $\beta$, the hit ratio and RPR of the recommended FSS algorithms vary as well. However, the change intervals fall in a relative small interval and the lower bound stands at a fairly high level. The minimum average hit ratio is up to 91.74% and the minimum average RPR is up to 97.68%. This indicates that the proposed FSS algorithm recommendation method has general application and works well for different settings of $\alpha$ and $\beta$.

## 5. Sensitivity Analysis of the Number of Nearest Data Sets on Recommendation Results

In this section, we analyze how the number of the nearest data sets affects the recommendation performance. Based on the experimental results, we provide some guidelines for selecting the appropriate number of nearest data sets in practice.

### 5.1 Experimental Method

Generally, different numbers of the nearest data sets (i.e., $k$) will result in different recommendations. Thus, when recommending FSS algorithms to a feature selection problem, an appropriate $k$ value is very important.

The $k$ value that results in higher recommendation performance is preferred. However, the recommendation performance difference under two different $k$ values sometimes might be random and not significant. Thus, in order to identify an appropriate $k$ value from alternatives, we should first determine whether or not the performance differences among them are statistically significant. Non-parametric statistical test, Friedman test followed by Holm procedure test as suggested by Demšar (2006), can be used for this purpose.

In the experiment, we conducted FSS algorithm recommendation with all possible $k$ values (i.e., from 1 to 114) over the 115 data sets. When identifying the appropriate $k$ values, the non-parametric statistical test is conducted as follows.

Firstly, the Friedman test is performed over the 114 recommendation performance at the significance level 0.05. Its null hypothesis is that the 114 $k$ values perform equivalently well in the proposed recommendation method over the 115 data sets.

If the Friedman test rejects the null hypothesis, that is, there exists significant difference among these 114 $k$ values, then we choose one under which the recommendation has the best performance as the reference. After that, the Holm procedure test is performed to find out the $k$ values under which the recommendation performance has no significant difference with that of the reference. The identified $k$ values including the reference are the appropriate numbers of the nearest data sets.

### 5.2 Results Analysis

Fig. 14 shows how the number of the nearest data sets (i.e., $k$) affects the performance of the recommendation method under different settings of $\alpha$ and $\beta$, where '×' denotes the $k$ under which the recommendation performance is significantly worse than others at the significance level of 0.05. From it we observe that:

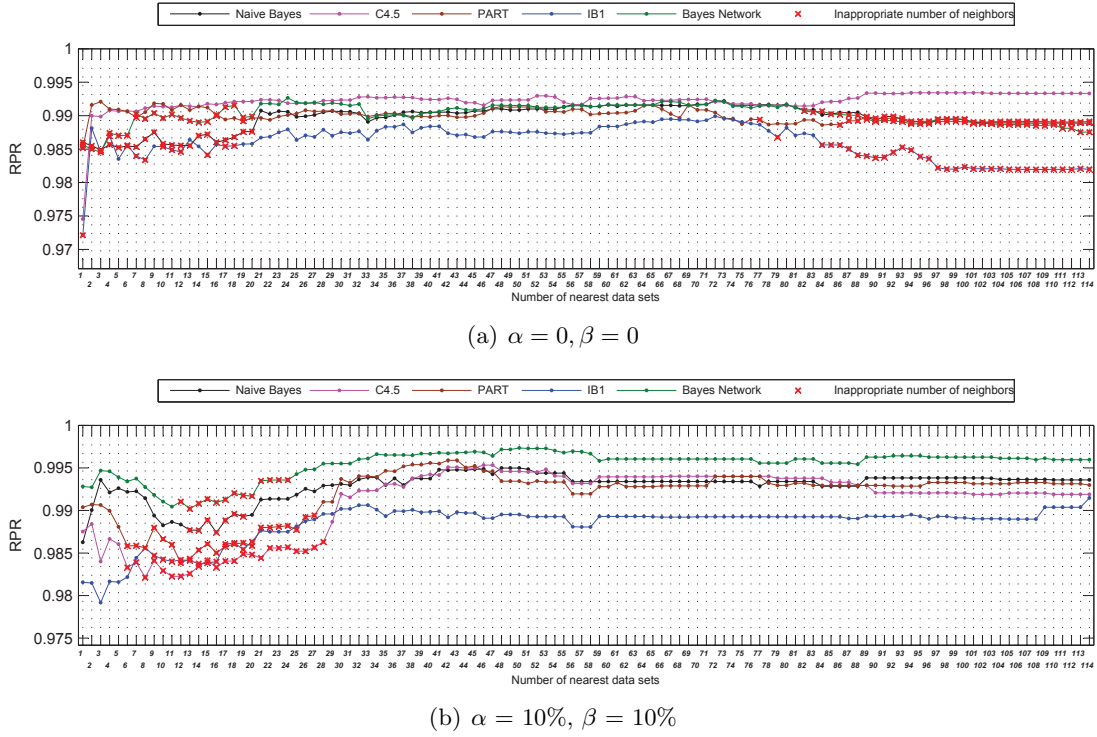

(a) $\alpha = 0, \beta = 0$

(b) $\alpha = 10\%, \beta = 10\%$

Figure 14: Number of the nearest data sets vs. $RPR$

1) When $\alpha = \beta = 0$ (Fig. 14(a)), for each of the five classifiers, the $RPR$ varies with different $k$ values. Specifically, the $RPR$ is fluctuant when $k$ is smaller than 20, while it is relatively flat in the middle part, and it decreases when $k$ is larger than 79 except for C4.5. However, the increment of C4.5 is very small ($< 0.002$). This might be due to that C4.5 picks up useful features to build the tree by itself, so the impact of other feature selection methods is less. Moreover, the difference among accuracies of C4.5 on most data sets is relatively small, while the performance metric $EARR$ that used to evaluate different FSS algorithms depends only on classification accuracy when $\alpha = \beta = 0$. Thus, the $RPR$ of C4.5 is relatively stable for the different values of $k$.

2) In the case of $\alpha = \beta = 10\%$ (Fig. 14(b)), the variation of $RPR$ is different from that of $\alpha = \beta = 0$. For each of the five classifiers, the $RPR$ first decreases with fluctuations, then increases, and finally decreases slowly and steadily. This could be due to that, when the parameters $\alpha$ and $\beta$ are set to be a relatively large value (such as 10% in our experiment), the runtime of ( or the number of features selected by) an FSS algorithm will play a more important role in evaluating the performance of the FSS algorithm. Thus, for a given data set, the FSS algorithms with lower time complexity (or the smaller number of selected features) will be more possibly higher ranked and have larger $RPR$. Therefore, with the increasing of $k$, these algorithms are more possibly recommended. Meanwhile, for most data sets, these algorithms are either the real appropriate algorithms or with larger $RPR$, so the $RPR$ averaged over all data sets is relatively stable with the increasing of $k$.

3) Comparing the cases of $\alpha = \beta = 0$ and $\alpha = \beta = 10\%$, we found that '×' appears when $k < 21$ for the former and $k < 29$ for the latter, while it emerges again when $k > 76$ for the former. This means we cannot choose the $k$ values falling into these ranges. At the same time, we also found that the peak values of $RPR$ for $\alpha = \beta = 10\%$ appear in the range of [32, 54], which is also one of the peak value ranges for $\alpha = \beta = 0$ except C4.5. This means if we set $k$ to 28% to 47% of the number of data sets, better recommendation performance can be obtained.

## 6. Conclusion

In this paper, we have presented an FSS algorithm recommendation method with the aim to support the automatic selection of appropriate FSS algorithms for a new feature selection problem from a number of candidates.

The proposed recommendation method consists of meta-knowledge database construction and algorithm recommendation. The former obtains the meta-features and the performance of all the candidate FSS algorithms, while the latter models the relationship between the meta-features and the FSS algorithm performance based on a $k$-NN method and recommends appropriate algorithms for a feature selection problem with the built up model.

We have thoroughly tested the recommendation method with 115 real world data sets, 22 different FSS algorithms, and five representative classification algorithms under two typical user's performance requirements. The experimental results show that our recommendation method is effective.

We have also conducted a sensitivity analysis to explore how the number of the nearest data sets ($k$) impacts the FSS algorithm recommendation, and suggest to set $k$ as the 28% to 47% of the number of the historical data sets.

In this paper, we have utilized the well-known and commonly-used meta-features to characterize different data sets. "Which meta-features are informative?" and "Are there any other more informative meta-features?" are still open questions. To our knowledge, there still does not exist any effective method to answer these questions. Thus, for future work, we plan to explore further that how to measure the information of the meta-features and whether there are some more informative meta-features that can lead to further improvements for FSS algorithm recommendation.

## Acknowledgements

This work is supported by the National Natural Science Foundation of China under grant 61070006.

## References

Aha, D. W., Kibler, D., & Albert, M. K. (1991). Instance-based learning algorithms. *Machine learning*, *6*(1), 37–66.

Ali, S., & Smith, K. A. (2006). On learning algorithm selection for classification. *Applied Soft Computing*, *6*(2), 119–138.

Atkeson, C. G., Moore, A. W., & Schaal, S. (1997). Locally weighted learning. *Artificial intelligence review, 11*(1), 11–73.

Battiti, R. (1994). Using mutual information for selecting features in supervised neural net learning. *IEEE Transactions on Neural Networks, 5*(4), 537–550.

Brazdil, P., Carrier, C., Soares, C., & Vilalta, R. (2008). *Metalearning: Applications to data mining.* Springer.

Brazdil, P. B., Soares, C., & Da Costa, J. P. (2003). Ranking learning algorithms: Using IBL and meta-learning on accuracy and time results. *Machine Learning, 50*(3), 251–277.

Brodley, C. E. (1993). Addressing the selective superiority problem: Automatic algorithm/model class selection. In *Proceedings of the Tenth International Conference on Machine Learning*, pp. 17–24. Citeseer.

Castiello, C., Castellano, G., & Fanelli, A. (2005). Meta-data: Characterization of input features for meta-learning. *Modeling Decisions for Artificial Intelligence*, 457–468.

Dash, M., & Liu, H. (1997). Feature selection for classification. *Intelligent data analysis, 1*(3), 131–156.

Dash, M., & Liu, H. (2003). Consistency-based search in feature selection. *Artificial Intelligence, 151*(1-2), 155–176.

de Souza, J. T. (2004). *Feature selection with a general hybrid algorithm.* Ph.D. thesis, University of Ottawa.

Demšar, J. (2006). Statistical comparisons of classifiers over multiple data sets. *Journal of Machine Learning Research, 7,* 1–30.

Engels, R., & Theusinger, C. (1998). Using a data metric for preprocessing advice for data mining applications..

Frank, E., & Witten, I. H. (1998). Generating accurate rule sets without global optimization. In *Proceedings of the 25th international conference on Machine learning*, pp. 144–151. Morgan Kaufmann, San Francisco, CA.

Friedman, M. (1937). The use of ranks to avoid the assumption of normality implicit in the analysis of variance. *Journal of the American Statistical Association, 32*(200), 675–701.

Friedman, N., Geiger, D., & Goldszmidt, M. (1997). Bayesian network classifiers. *Machine learning, 29*(2), 131–163.

Gama, J., & Brazdil, P. (1995). Characterization of classification algorithms. *Progress in Artificial Intelligence*, 189–200.

Garci'a Lopez, F., Garci'a Torres, M., Melian Batista, B., Moreno Perez, J. A., & Moreno-Vega, J. M. (2006). Solving feature subset selection problem by a parallel scatter search. *European Journal of Operational Research, 169*(2), 477–489.

Goldberg, D. E. (1989). *Genetic algorithms in search, optimization, and machine learning.* Addison-Wesley Professional.

Gutlein, M., Frank, E., Hall, M., & Karwath, A. (2009). Large-scale attribute selection using wrappers. In *Proceedings of IEEE Symposium on Computational Intelligence and Data Mining*, pp. 332–339. IEEE.

Guyon, I., & Elisseeff, A. (2003). An introduction to variable and feature selection. *The Journal of Machine Learning Research*, *3*, 1157–1182.

Hall, M., Frank, E., Holmes, G., Pfahringer, B., Reutemann, P., & Witten, I. (2009). The weka data mining software: an update. *ACM SIGKDD Explorations Newsletter*, *11*(1), 10–18.

Hall, M. A. (1999). *Correlation-based Feature Selection for Machine Learning*. Ph.D. thesis, The University of Waikato.

Hedar, A. R., Wang, J., & Fukushima, M. (2008). Tabu search for attribute reduction in rough set theory. *Soft Computing-A Fusion of Foundations, Methodologies and Applications*, *12*(9), 909–918.

Hommel, G. (1988). A stagewise rejective multiple test procedure based on a modified bonferroni test. *Biometrika*, *75*(2), 383–386.

John, G. H., & Langley, P. (1995). Estimating continuous distributions in Bayesian classifiers. In *Proceedings of the eleventh conference on uncertainty in artificial intelligence*, Vol. 1, pp. 338–345. Citeseer.

Kalousis, A., Gama, J., & Hilario, M. (2004). On data and algorithms: Understanding inductive performance. *Machine Learning*, *54*(3), 275–312.

King, R. D., Feng, C., & Sutherland, A. (1995). Statlog: comparison of classification algorithms on large real-world problems. *Applied Artificial Intelligence*, *9*(3), 289–333.

Kira, K., & Rendell, L. (1992). A practical approach to feature selection. In *Proceedings of the ninth international workshop on Machine learning*, pp. 249–256. Morgan Kaufmann Publishers Inc.

Kohavi, R. (1995). A study of cross-validation and bootstrap for accuracy estimation and model selection. In *International joint Conference on artificial intelligence*, Vol. 14, pp. 1137–1145. Citeseer.

Kohavi, R., & John, G. (1997). Wrappers for feature subset selection. *Artificial intelligence*, *97*(1), 273–324.

Kononenko, I. (1994). Estimating attributes: analysis and extensions of RELIEF. In *Proceedings of the European conference on machine learning on Machine Learning*, pp. 171–182. Springer-Verlag New York.

Lee, M., Lu, H., Ling, T., & Ko, Y. (1999). Cleansing data for mining and warehousing. In *Proceedings of the 10th International Conference on Database and Expert Systems Applications*, pp. 751–760. Springer.

Lindner, G., & Studer, R. (1999). AST: Support for algorithm selection with a CBR approach. *Principles of Data Mining and Knowledge Discovery*, 418–423.

Liu, H., Motoda, H., Setiono, R., & Zhao, Z. (2010). Feature Selection: An Ever Evolving Frontier in Data Mining. In *The Fourth Workshop on Feature Selection in Data Mining*, pp. 3–14. Citeseer.

Liu, H., & Setiono, R. (1995). Chi2: Feature selection and discretization of numeric attributes. In *Proceedings of the Seventh International Conference on Tools with Artificial Intelligence*, pp. 388–391. IEEE.

Liu, H., & Setiono, R. (1996). A probabilistic approach to feature selection-a filter solution.. pp. 319–327. Citeseer.

Liu, H., & Yu, L. (2005). Toward integrating feature selection algorithms for classification and clustering. *IEEE Transactions on Knowledge and Data Engineering*, *17*(4), 491–502.

Michie, D., Spiegelhalter, D. J., & Taylor, C. C. (1994). Machine learning, neural and statistical classification..

Molina, L. C., Belanche, L., & Nebot, À. (2002). Feature selection algorithms: A survey and experimental evaluation. In *Proceedings of IEEE International Conference on Data Mining*, pp. 306–313. IEEE.

Nakhaeizadeh, G., & Schnabl, A. (1997). Development of multi-criteria metrics for evaluation of data mining algorithms. In *Proceedings of the 3rd International Conference on Knowledge Discovery and Data mining*, pp. 37–42.

Nakhaeizadeh, G., & Schnabl, A. (1998). Towards the personalization of algorithms evaluation in data mining. In *Proceedings of the 4th International Conference on Knowledge Discovery and Data mining*, pp. 289–293.

Pudil, P., Novovičová, J., & Kittler, J. (1994). Floating search methods in feature selection. *Pattern recognition letters*, *15*(11), 1119–1125.

Pudil, P., Novovičová, J., Somol, P., & Vrňata, R. (1998a). Conceptual base of feature selection consulting system. *Kybernetika*, *34*(4), 451–460.

Pudil, P., Novovičovà, J., Somol, P., & Vrňata, R. (1998b). Feature selection expertuser oriented approach. *Advances in Pattern Recognition*, 573–582.

Quinlan, J. R. (1993). *C4.5: programs for machine learning*. Morgan Kaufmann.

Robnik-Šikonja, M., & Kononenko, I. (2003). Theoretical and empirical analysis of relieff and rrelieff. *Machine learning*, *53*(1), 23–69.

Saeys, Y., Inza, I., & Larrañaga, P. (2007). A review of feature selection techniques in bioinformatics. *Bioinformatics*, *23*(19), 2507–2517.

Smith-Miles, K. A. (2008). Cross-disciplinary perspectives on meta-learning for algorithm selection. *ACM Computing Surveys*, *41*(1), 1–25.

Sohn, S. Y. (1999). Meta analysis of classification algorithms for pattern recognition. *IEEE Transactions on Pattern Analysis and Machine Intelligence*, *21*(11), 1137–1144.

Song, Q. B., Wang, G. T., & Wang, C. (2012). Automatic recommendation of classification algorithms based on data set characteristics. *Pattern Recognition*, *45*(7), 2672–2689.

Vilalta, R., & Drissi, Y. (2002). A perspective view and survey of meta-learning. *Artificial Intelligence Review*, *18*(2), 77–95.

Wolpert, D. H. (2001). The supervised learning no-free-lunch theorems. In *Proceedings of 6th Online World Conference on Soft Computing in Industrial Applications*, pp. 25–42. Citeseer.

Yu, L., & Liu, H. (2003). Feature selection for high-dimensional data: A fast correlation-based filter solution. In *Proceedings of The Twentieth International Conference on Machine Leaning*, Vol. 20, pp. 856–863.

Zhao, Z., & Liu, H. (2007). Searching for interacting features. In *Proceedings of the 20th International Joint Conference on Artifical Intelligence*, pp. 1156–1161. Morgan Kaufmann Publishers Inc.

Zhou, X., & Dillon, T. (1988). A heuristic-statistical feature selection criterion for inductive machine learning in the real world. In *Proceedings of the IEEE International Conference on Systems, Man, and Cybernetics*, Vol. 1, pp. 548–552. IEEE.